\documentclass[journal]{IEEEtran}

\usepackage{graphicx}
\usepackage{hyperref}

\usepackage[dvipsnames]{xcolor}
\colorlet{LightRubineRed}{black!100!}
\colorlet{Mycolor1}{green!5!orange!95!}
\definecolor{Mycolor2}{HTML}{00F9DE}
\definecolor{mygray}{gray}{1.0}

\begin{document}

%\begin{frontmatter}

\title{A Bottom-up Approach for Pancreas Segmentation using Cascaded Superpixels and (Deep) Image Patch Labeling}

\author{Amal Farag, Le Lu, {\em Senior Member, IEEE}, Holger R. Roth, Jiamin Liu, Evrim Turkbey, Ronald M. Summers 
\thanks{This work was performed when all authors were at Department of Radiology and Imaging Sciences, National Institutes of Health Clinical Center, Bethesda, MD 20892-1182, USA. Amal Farag is currently affiliated with Kentucky Imaging and Technologies, Louisville, Kentucky; and Evrim Turkbey is with School of Medicine, Johns Hopkins University, 733 N Broadway, Baltimore, Maryland. e-mail: \{le.lu, rms\}@nih.gov.}
}

\maketitle

\begin{abstract}
Robust automated organ segmentation is a prerequisite for computer-aided diagnosis (CAD), quantitative imaging analysis, detection of pathologies and surgical assistance. For anatomical high-variability organs such as the pancreas, previous segmentation approaches report low accuracies in comparison to well studied organs like the liver or heart. We present a fully-automated bottom-up approach for pancreas segmentation in abdominal computed tomography (CT) scans. The method is based on a hierarchical cascade of information propagation by classifying image patches at different resolutions and cascading (segments) superpixels. There are four stages in the system: 1) decomposing CT slice images as a set of disjoint boundary-preserving superpixels; 2) computing pancreas class probability maps via dense patch labeling; 3) classifying superpixels by pooling both intensity and probability features to form empirical statistics in cascaded random forest frameworks; and 4) simple connectivity based post-processing. The dense image patch labeling is conducted by two schemes: efficient random forest classifier on image histogram, location and texture features; and more expensive (but with better specificity) deep convolutional neural network classification, on larger image windows (i.e., with more spatial contexts). Over-segmented $2D$ CT slices by the Simple Linear Iterative Clustering approach are adopted through model/parameter calibration and labeled at the superpxiel level for positive (pancreas) or negative (non-pancreas background) classes. %A supervised random forest (RF) classifier is trained on the patch level and at the superpixel level, coupled with multi-channel feature extraction.
Evaluation of the approach is done on a database of $80$ manually segmented CT volumes in six-fold cross-validation. Our achieved results are comparable, or better than the state-of-the-art methods (evaluated by ``leave-one-patient-out''), with a Dice coefficient of $70.7\%$ and Jaccard Index of $57.9\%$. In addition, the computational efficiency has been drastically improved in the order of $6\sim8$ minutes, comparing with others of $\geq 10$ hours per testing case. The segmentation framework using deep patch labeling confidences is also more numerically stable, reflected by the smaller performance metric standard deviations. \textcolor{LightRubineRed} {Finally, we implement a multi-atlas label fusion (MALF) approach for pancreas segmentation using the same dataset. Under six-fold cross-validation, our bottom-up segmentation method significantly outperforms its MALF counterpart: $70.7 \pm 13.0\%$ versus $52.51 \pm 20.84\%$ in Dice coefficients.}
\end{abstract}

%\begin{keyword}
%Abdominal Computed Tomography (CT), Superpixels, Image Patch Labeling, Deep Convolutional Neural Networks, Patch-level Confidence Pooling, Cascaded Random Forest, Pancreas Segmentation.
%\end{keyword}

%\end{frontmatter}

%\linenumbers

\section{Introduction}
Image segmentation is a key step in image understanding that aims at separating objects within an image into classes, based on object characteristics and a prior information about the surroundings. This also applies to medical image analysis in various imaging modalities. The segmentation of abdominal organs such as the spleen, liver and pancreas in abdominal computed tomography (CT) scans can be an important input to computer aided diagnosis (CAD) systems, for quantitative and qualitative analysis and for surgical assistance. In the instance of quantitative imaging analysis of diabetic patients, a requisite critical step for the development of such CAD systems is segmentation specifically of the pancreas. Pancreas segmentation is also a necessary input for subsequent methodologies for pancreatic cancer detection. The literature is rich in methods of automatic segmentation on CT with high accuracies (e.g., Dice coefficients $>90\%$), of other organs such as the kidneys \cite{Cuingnet12}, lungs \cite{Mansoor14}, heart \cite{Zheng08} and liver \cite{Ling08}. Yet, high accuracy in automatic segmentation of the pancreas remains a challenge. The literature is not as abundant in either single- or multi-organ segmentation setups.

\begin{figure}[t]
\begin{center}
\includegraphics[width=0.80\linewidth]{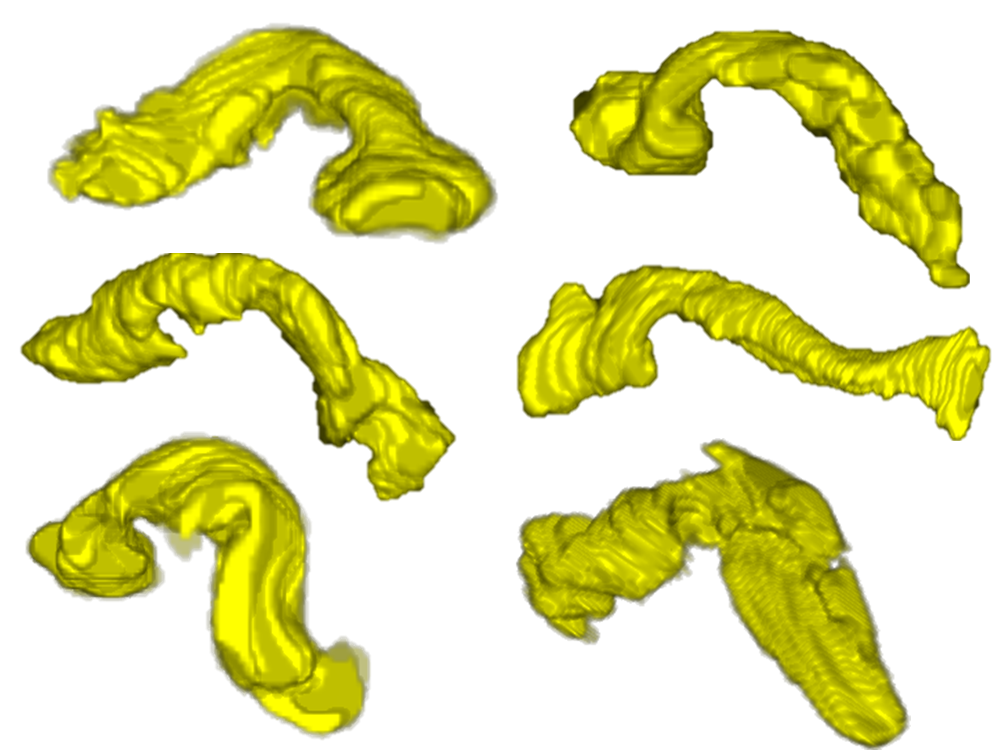}\\
\end{center}
\caption{3D manually segmented volumes of six pancreases from six patients.Notice the shape and size variations.}\label{fig:img1}
\end{figure}

The pancreas is a highly anatomically variable organ in terms of shape and size and the location within the abdominal cavity shifts from patient to patient. The boundary contrast can vary greatly by the amount of visceral fat in the proximity of the pancreas. These factors and others make segmentation of the pancreas very challenging. Fig. \ref{fig:img1} depicts several manually segmented 3D volumes of various patient pancreases to better illustrate the variations and challenges mentioned. From the above observations, we argue that the automated pancreas segmentation problem should be treated differently, apart from the current organ segmentation literature where statistical shape models are generally used.

%%%%%%%%%%

In this paper, a new fully bottom-up approach using image and (deep) patch-level labeling confidences for pancreas segmentation is proposed using 80 single phase CT patient data volumes. The approach is motivated to improve the segmentation accuracy of {\em highly deformable organs}, like the pancreas, by leveraging {\em middle-level representation of image segments}. First, over segmentation of all 2D slices of an input patient abdominal CT scan is obtained as a semi-structured representation known as superpixels. Second, classifying superpixels into two semantic classes of pancreas and non-pancreas is conducted as a multi-stage feature extraction and random forest (RF) classification process, on the image and (deep) patch-level confidence maps, pooled at the superpixel level. Two cascaded random forest superpixel classification frameworks are addressed and compared. Fig. \ref{fig:Framework} depicts the overall proposed first framework. Fig. \ref{fig:Classifier_Flow} illustrates the modularized flow charts of both frameworks. Our experimental results are carried-out in a six-fold cross-validation manner. Our system runs at about two orders of magnitude more computationally efficiently to process a new testing case than the atlas registration based approaches \cite{Shimizu,Okada,Wolz,Chu,Wolz12,Wang}. The obtained results are comparable, or better than the state-of-the-art methods (evaluated by ''leave-one-patient-out''), with a Dice coefficient of $70.7\%$ and Jaccard Index of $57.9\%$. \textcolor{LightRubineRed} {Under the same six-fold cross-validation, our bottom-up segmentation method significantly outperforms its ``multi-atlas registration and joint label fusion'' (MALF) counterpart (based on our implementation using \cite{Modat2010,Wang2012}): Dice coefficients $70.7 \pm 13.0\%$ versus $52.51 \pm 20.84\%$.  Additionally, another bottom-up supervoxel based multi-organ segmentation without registration in 3D abdominal CT images is also investigated \cite{Zografos2015} in a similar spirit, for demonstrating this methodological synergy.}

\section{Previous Work}

The organ segmentation literature can be divided into two broad categories: top-down and bottom-up approaches. In top-down approaches, a-priori knowledge such as atlas(es) and/or shape models of the organ are generated and incorporated into the framework via learning based shape model fitting \cite{Cuingnet12,Zheng08,Ling08} or volumetric image registration \cite{Wolz,Chu,Wang}. For bottom-up approaches segmentation are performed by local image similarity grouping and growing \cite{Mansoor14} or pixel, superpixel/supervoxel based labeling \cite{Lucchi} since direct representations of the organ is not incorporated. Generally speaking, top-down methods are targeted for organs which can be modeled well by statistical shape models \cite{Cuingnet12,Zheng08} whereas bottom-up representations are more effective for highly non-Gaussian shaped \cite{Lucchi} or pathological organs.

Previous work on pancreas segmentation from CT images have been dominated by top-down approaches which rely on atlas based approaches or statistical shape modeling or both \cite{Shimizu,Okada,Wolz,Chu,Wolz12,Wang}. \cite{Shimizu} is generally not comparable to others since it uses three-phase contrast enhanced CT datasets. The rest are performed on single phase CT images. %``Leave-one-patient-out'' (LOO) cross-validation metric is adopted for all \cite{Shimizu,Okada,Wolz,Chu,Wolz12,Wang}.

\begin{itemize}

\item Shimizu et. al \cite{Shimizu} utilize three-phase contrast enhanced CT data which are first registered together for a particular patient and then registered to a reference patient by landmark-based deformable registration. The spatial support area of the abdominal cavity is reduced by segmenting the liver, spleen and three main vessels associated with location interpretation of the pancreas (i.e. splenic, portal and superior mesenteric veins). Coarse-to-fine pancreas segmentation is performed by using generated patient-specific probabilistic atlas guided segmentation followed by intensity-based classification and post-processing. Validation of the approach was conducted on 20 multi-phase datasets resulting in a Jaccard of $57.9\%$.

\item Okada et. al \cite{Okada} perform multi-organ segmentation by combining inter-organ spatial interrelations with probabilistic atlases. The approach incorporated various a-priori knowledge into the model that includes shape representations of seven organs. Experimental validation was conducted on 28 abdominal contrast-enhanced CT datasets obtaining an overall volume overlap of Dice index 46.6\% for the pancreas.

\item Chu et. al \cite{Chu} present an automated multi-organ segmentation method based on spatially-divided probabilistic atlases. The algorithm consists of image-space division and a multi-scale weighting scheme to deal with the large differences among patients in organ shape and position in local areas. Their experimental results show that the liver, spleen, pancreas and kidneys can be segmented with Dice similarity indices of 95.1\%, 91.4\%, 69.1\%, and 90.1\%, respectively, using 100 annotated abdominal CT volumes.

\item Wolz et. al \cite{Wolz} may be considered the state-of-the-art result thus far for single-phase pancreas segmentation. The approach is a multi-organ segmentation approach that combines hierarchical weighted subject-specific atlas-based registration and patch-based segmentation. Post-processing is in the form of optimized graph-cuts with a learned intensity model. Their results in terms of a Dice overlap for the pancreas is 69.6\% on 150 patients and 58.2\% on a sub-population of 50 patients.

\item Recent work by Wang et. al \cite{Wang} proposes a patch-based label propagation approach that uses relative geodesic distances. The approach can be considered a start to developing some bottom-up component for segmentation, where affine registration between dataset and atlases were conducted followed by refinement using the patch-based segmentation to reduce misregistrations and instances of high anatomy variability. The approach was evaluated on 100 abdominal CT scans with an overall Dice of 65.5\% for the pancreas segmentation.

\end{itemize}

\textcolor{LightRubineRed}
{The default experimental setting in many of the atlas based approaches \cite{Shimizu,Okada,Wolz,Chu,Wolz12,Wang} is conducted in a ``leave-one-patient-out'' or ``leave-one-out'' (LOO) criterion for up to N=150 patients. In the clinical setting leave-one-out based dense volume registration (from all other N-1 patients as atlas templates) and label fusion process may be computationally impractical (10+ hours per testing case). More importantly, it does not scale up easily when large scale datasets are present.}

%%%%%%%
\begin{figure}[t]
\begin{center}
\includegraphics[width=0.96\linewidth]{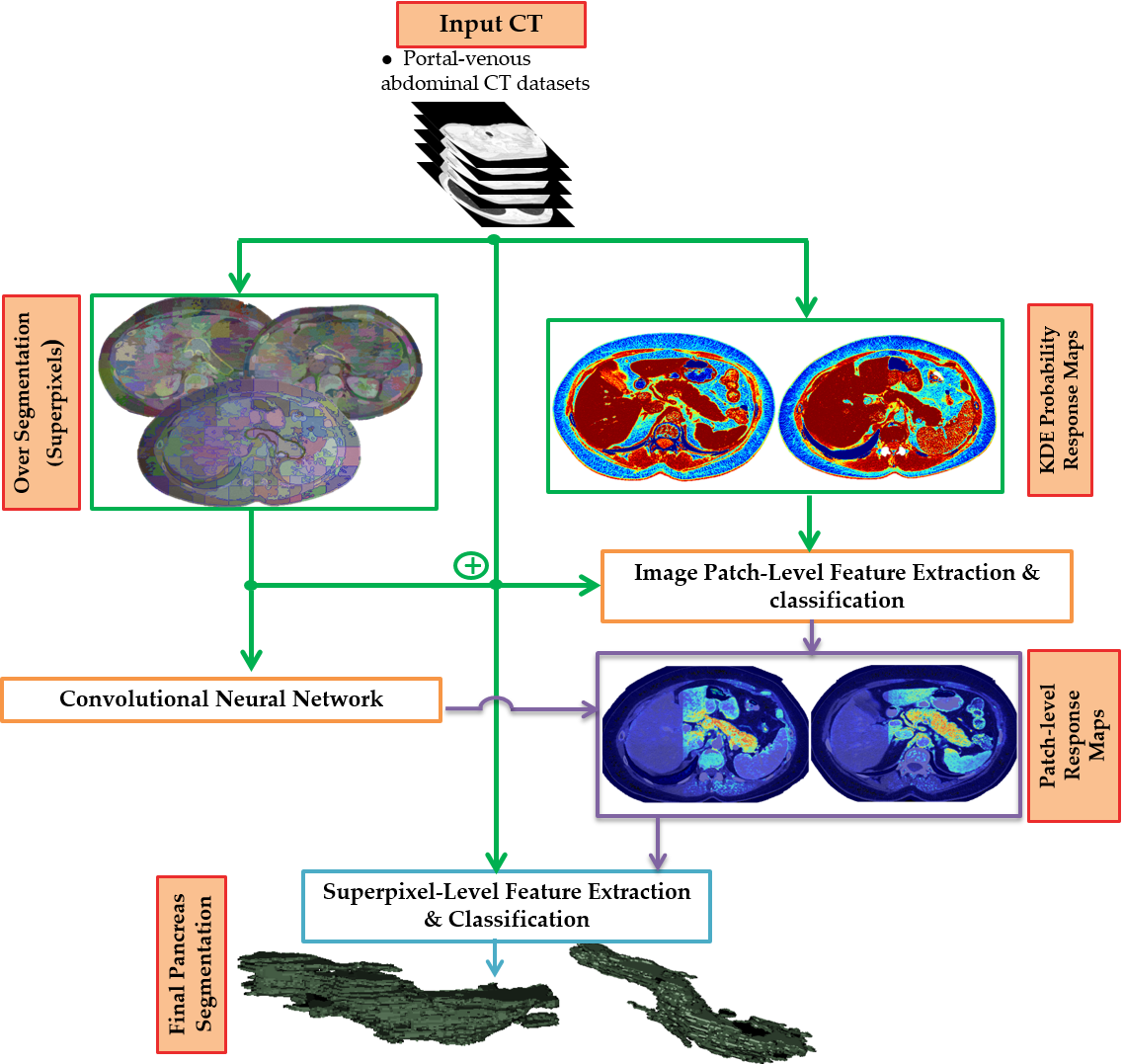}\\
\end{center}
\caption{Overall pancreas segmentation framework via dense image patch labeling.}\label{fig:Framework}
\end{figure}
%%%%%%%%%%%%

%The default experimental setting in many of the atlas based approaches \cite{Wang,Wolz,Chu,Wolz12,Shimizu} is conducted in a ``leave-one-patient-out'' or ``leave-one-out'' criterion for up to N=150 patients. In the clinical setting leave-one-out based dense volume registration (from all other N-1 patients as atlas templates) and label fusion process may be computationally impractical (10+ hours per testing case). More importantly, it does not scale up easily when large scale datasets are present. As such, we employed 6-fold cross-validation. The information learned from the training folds is captured and compactly encoded as the image patch-level and superpixel-level classifiers (RFs and CNN).

The proposed bottom-up approach is significantly more efficient in memory and computation speed than the multi-atlas registration framework \cite{Wang,Wolz,Chu,Wolz12,Shimizu,Okada}. Quantitative evaluation on $80$ manually segmented CT patient volumes under six-fold cross-validation (CV) are conducted. Our results are comparable, or better than the state-of-the-art methods (even under ``leave-one-patient-out'', or LOO), with a Dice coefficient $70.7\%$ and Jaccard Index $57.9\%$. The strict numerical performance comparison is not possible since experiments can not be performed on the same datasets\footnote{Our annotated pancreas segmentation datasets are publicly available at https://wiki.cancerimagingarchive.net/display/Public/Pancreas-CT, to ease future comparisons.}. We instead implement a multi-atlas label fusion (MALF) pancreas segmentation approach \cite{Modat2010,Wang2012} using our datasets. Under six-fold CV, our bottom-up segmentation method clearly outperforms its MALF counterpart: Dice coefficients $70.7 \pm 13.0\%$ versus $52.51 \pm 20.84\%$.

Superpixel based representation on pathological region detection and segmentation is recently studied in \cite{Mahapatra} using MRI datasets. However the problem representation, visual feature extraction and classification framework \cite{Mahapatra} vary drastically from our approach. Our bottom-up image parsing method, especially using superpixels as an intermediate level image representation is partially inspired by similar approaches in PASCAL semantic segmentation \cite{Everingham2015} and scene labeling \cite{Tighe2010,Tighe2013}. The technical and algorithm details are significantly different from \cite{Everingham2015,Tighe2010,Tighe2013}, especially in the sense that the pancreas is a relatively small organ and the segmentation areas or volumes of foreground/background (i.e., pancreas versus other) classes are highly unbalanced. Pancreas normally only consumes less than $1\%$ of space given an input CT scan. Here we use ``Cascaded Superpixels'' representation as a solution to address this domain-specific challenge that may widely exist in medical image analysis.

\textcolor{LightRubineRed}
%{A preliminary version of this work appears }. %A preliminary version of the approach has been published in \cite{Farag}. In this article an extension of using deep convolutional neural networks (CNN) to obtain and aggregate the dense patch labeling probabilities has been exploited, in conjunction with the image patch-level labels, to leverage rich CNN image features.
In this paper, we have generalized the proposed algorithm framework (Fig. \ref{fig:Framework}) in \cite{Farag2014}, as a preliminary version. From the second proposed framework (F-2) in Fig. \ref{fig:Classifier_Flow}, deep convolutional neural networks (CNN) based image patch labeling (to leverage richer CNN image features) is integrated and thoroughly evaluated (Sec. \ref{subsec:CNN} and Sec. \ref{subsec:Seg} are added as new content). More importantly, the representation aspects of different algorithm variations and experimental configurations; quantitative comparison to MALF \cite{Modat2010,Wang2012} and superpixel based CNNs \cite{Roth2015,Girshick2015} under 6 or 4-fold CV, and technical insights are extensively enriched to improve the completeness of this manuscript.

\section{Methods}

In this section the components of our overall algorithm flow (shown in Fig. \ref{fig:Framework}) is first addressed (Sec. \ref{subsec:SP} and Sec. \ref{subsec:patch}). The method extensions on exploiting sliding-window CNN based dense image patch labeling and framework variations are described in Sec. \ref{subsec:CNN} and Sec. \ref{subsec:Seg}.

\subsection{Boundary-preserving over-segmentation}\label{subsec:SP}

Over-segmentation occurs when images (or more generally grid graphs) are segmented or decomposed into smaller perceptually meaningful regions, ``superpixels''. Within a superpixel, pixels carry similarities in color, texture, intensity, etc. and generally align with image edges rather than rectangular patches (i.e. superpixels can be irregular in shape and size). In the computer vision literature, numerous approaches have been proposed for superpixel segmentation \cite{Achanta,Neubert,Liu,Felzenszwalb,Vincent}. Each approach has its drawbacks and advantages but three main properties are generally examined when deciding the appropriate method for an application as discussed in \cite{Neubert}: 1) adherence to image boundaries; 2) computationally fast, ease of usage and memory efficient; especially when computational complexity reduction is of importance and 3) improvement on both quality and speed of the final segmentation output.

\begin{figure}[t]
\begin{center}
\includegraphics[width=0.92\linewidth]{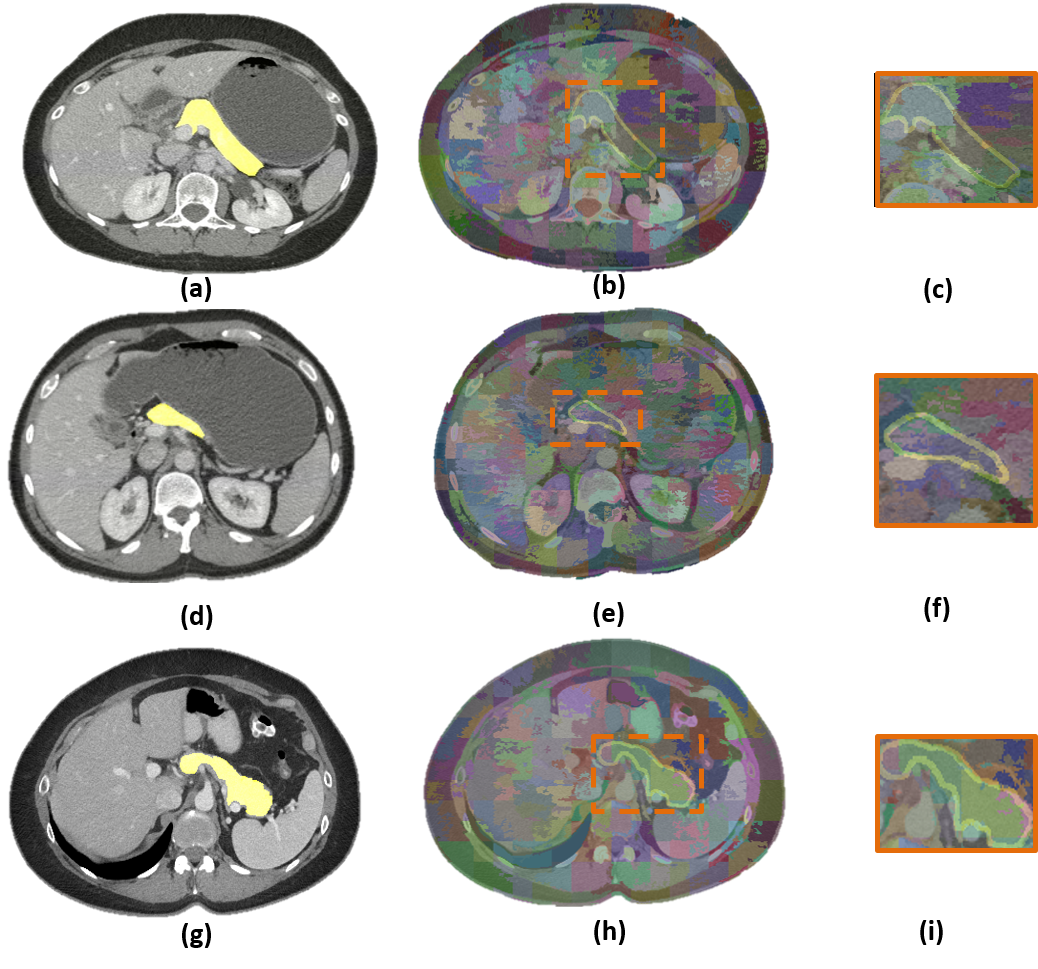}\\
\end{center}
\caption{Sample superpixel generation results from the SLIC method \cite{Achanta}. First column depicts different slices from different patient scans with the ground-truth pancreas segmentation in yellow (a, d \& g). The second column depicts the over segmentation results with the pancreas contours superimposed on the image (b, e \& h). Last, (c) (f) and (i) show zoomed-in areas of the pancreas superpixel results from (b) (e) and (h).}\label{fig:SLICRes}
\end{figure}

\begin{figure}[t]
\begin{center}
\includegraphics[width=0.68\linewidth]{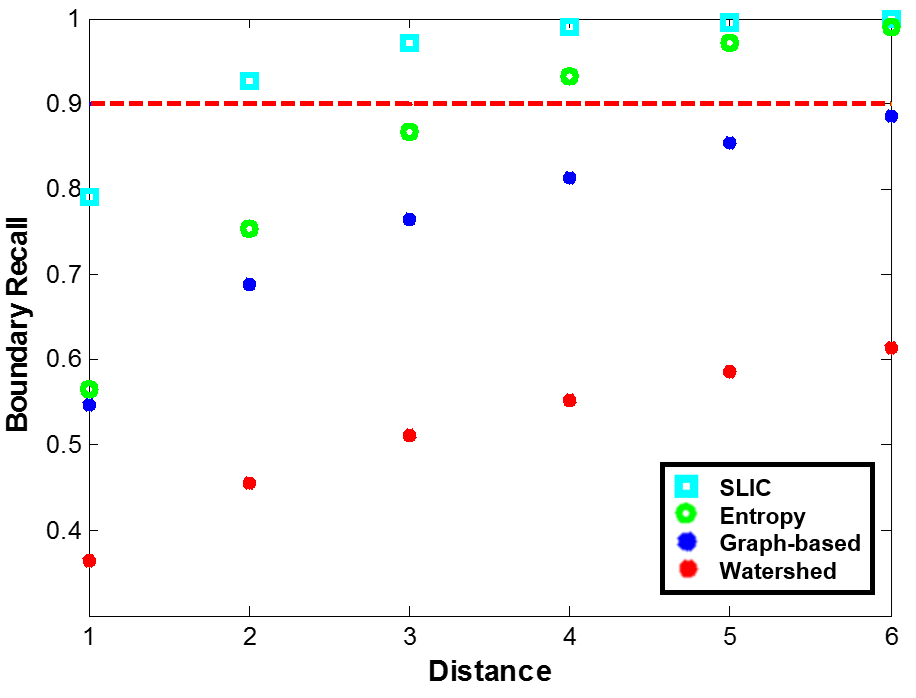}\\
\end{center}
\caption{Superpixels boundary recall results evaluated on 20 patient scans (Distance in millimeters). The watershed method \cite{Vincent} is shown in red, efficient graph \cite{Felzenszwalb} in blue while the SLIC \cite{Achanta} and the Entropy rate \cite{Liu} based methods are depicted in cyan and green, respectively. The red line represents the 90\% marker.}\label{fig:boundRec}
\end{figure}

Superpixel methods fall under two main broad categories: graph-based (e.g. SLIC \cite{Achanta}, entropy rate \cite{Liu} and \cite{Felzenszwalb}) and gradient ascent methods (e.g. watershed \cite{Vincent} and mean shift \cite{Comaniciu02}). In terms of computational complexity, \cite{Felzenszwalb,Vincent} are relatively fast in $O(MlogM)$ complexity where $M$ is the number of pixels or voxels in the image or grid graph. Mean shift \cite{Comaniciu02} and normalized cut \cite{Cour05} are $O(M^2)$, or $O(M^{\frac{3}{2}})$, respectively. Simple linear iterative clustering (SLIC) \cite{Achanta} is both fast and memory efficient. In our work, evaluation and comparison among three graph-based superpixel algorithms (i.e. SLIC \cite{Achanta,Neubert}, efficient graph-based \cite{Felzenszwalb} and Entropy rate \cite{Liu}) and one gradient ascent method (i.e. watershed \cite{Vincent}) are conducted, considering the three criterion in \cite{Neubert}. Fig. \ref{fig:SLICRes} shows sample superpixel results using the SLIC approach. The original CT slices and cropped zoomed-in pancreas superpixel regions are demonstrated. The boundary recall, a typical measurement used in the literature, to indicate how many ``true'' edge pixels of the ground truth object segmentation are within a pixel range from the superpixels (i.e., object-level edges are recalled by superpixel boundaries). High boundary recall indicates minimal true edges were neglected. Fig. \ref{fig:boundRec} shows sample quantitative results. Based on Fig. \ref{fig:SLICRes}, high boundary recalls, within the distance ranges between 1 and 6 pixels from the semantic pancreas ground-truth boundary annotation are obtained using the SLIC approach. The watershed approach provided the least promising results for usage in the pancreas, due to the lack of conditions in the approach, to utilize boundary information in conjunction with intensity information as implemented in graph-based approaches. The superpixel number range per axial image is constrained $\in [100,200]$ to make a good trade-off on superpixel dimensions or sizes.

%%%%%%%%%%%%%%
%%%%%%%%%%%%

The overlapping ratio $r$ of the superpixel versus the ground-truth pancreas annotation mask is defined as the percentage of pixels/voxels inside each superpixel  that are annotated as pancreas. By thresholding on $r$, say if $r>\tau$ the superpixel will be labeled as pancreas and otherwise as background, we can obtain the pancreas segmentation results. When $\tau=0.50$, the achieved mean Dice coefficient is $81.2\%\pm3.3\%$ which is referred as the ``Oracle'' segmentation accuracy since computing $r$ would require to know the ground-truth segmentation. This is also the upper bound segmentation accuracy for our superpixel labeling or classification framework. $81.2\pm3.3\%$ is significantly higher and numerically more stable (in standard deviation) than previous state-of-the-art methods \cite{Wang,Wolz,Chu,Wolz12,Shimizu}, to provide considerable improvement space of our work. Note that both the choices of SLIC and $\tau=0.50$ are calibrated using a subset of 20 scans. We find there is no need to evaluate different superpixel generation methods/parameters and $\tau$s as ``model selection'' using the training folds in each round of six-fold cross-validation. This superpixel calibration procedure is generalized well to all our datasets. Voxel-level pancreas segmentation can be propagated from superpixel-level classification and further improved by efficient narrow-band level-set based curve evolution \cite{Shi08}, or the learned intensity model based graph-cut \cite{Wolz}.

%%%%%%%%%%%%%
%%%%%%%%%%%%

Mid-level visual representation like superpixels has been widely adopted in recent {\em PASCAL VOC} semantic segmentation challenges \cite{Everingham2015} from computer vision. Generally speaking, a diverse set of plausible segments are expected per image to be fully exploited by later employed probabilistic models \cite{Yadollahpour}. Particularly, CPMC (Constrained Parametric Min-Cuts) based image segment generation process has achieved the state-of-the-art results for 20 object categories \cite{Carreira12a}. Segments are extracted around regularly placed foreground seeds against varying background seeds (with respect to to image boundary edges) using all levels of foreground bias \cite{Carreira12a}. CPMC has the effect of producing diverse segmentations at various locations and spatial scales however it is computationally expensive (in the order of 10 minutes for a natural image of 256$\times$256 pixels where our CT slices are at the resolution of 512$\times$512). For many patients (especially those with low body fat indexes) in our study, the contrast strengths of pancreas boundaries can be much weaker than PASCAL images (where the challenge more lies on the visually cluttered nature of multi-class objects). CPMC may not be effective on preserving pancreas boundaries.

Extension of superpixels to supervoxels is possible but in this work preference is made to 2D superpixel representation, due to the potential boundary leakage problem of supervoxels that may deteriorate the pancreas segmentation severely in multiple CT slices. Object level pancreas boundaries in abdominal CT images can appear very weakly, especially for patients with less body fat. On the other hand, as in Sec. \ref{subsec:patch}, image patch based appearance descriptors and statistical models are well developed for 2D images (in both computer vision and medical imaging). Using 3D statistical texture descriptors and models to  accommodate supervoxels can be computationally expensive, but without obvious performance improvement \cite{Lucchi,Arbelaez}.

In the next Sec. \ref{subsec:patch} and Sec. \ref{subsec:CNN}, we describe two different methods for dense image patch labeling using histogram/texture features and a deep Convolutional Neural Network.

\subsection{Patch-level Visual Feature Extraction and Classification: $P^{RF}$} \label{subsec:patch}

Feature extraction is a form of object representation that aims at capturing the important shape, texture and other salient features that allow distinctions between the desired object (i.e. pancreas) and the surrounding to be made. In this work a total of $46$ patch-level image features to depict the pancreas and its surroundings are implemented. The overall 3D abdominal body region per patient is first segmented and identified using a standard table-removal procedure where all voxels outside the body are removed.

\begin{figure}[t]
\begin{center}
\includegraphics[width=0.82\linewidth]{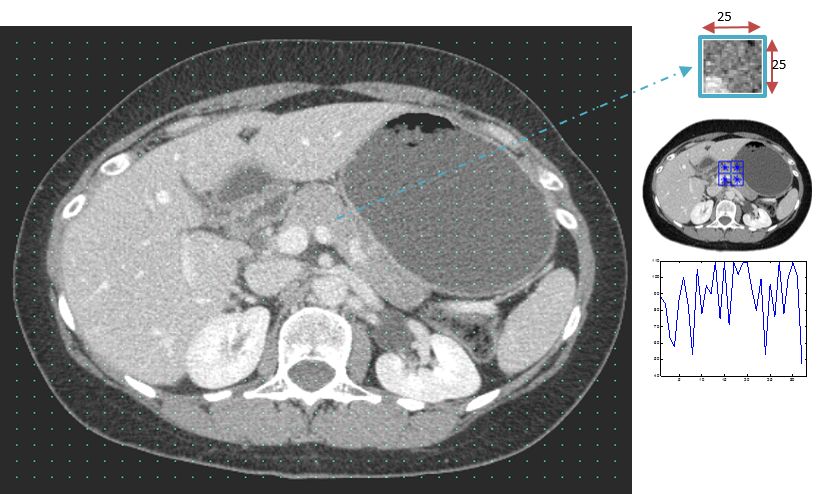}
\end{center}
\caption{Sample slice with center positions superimposed as green dots. The 25$\times$25 image patch and corresponding D-SIFT descriptors are shown to the right of the original image.}\label{fig:dsift}
\end{figure}

1), To describe the texture information, we adopt the Dense Scale-Invariant Feature transform (dSIFT) approach \cite{dsift} which is derived from the SIFT descriptor \cite{sift} with several technical extensions. The publicly available VLFeat implementation of the dSIFT is employed \cite{dsift}. Fig. \ref{fig:dsift} depicts the process implemented on a sample image slice. The descriptors are densely and uniformly extracted from image grids with inter-distances of 3 pixels. The patch center position are shown as the green points superimposed on the original image slice. Once the positions are known, the dSIFT is computed with the geometry of $\left[2x2\right]$ bins and bin size of 6 pixels, which results in a 32 dimensional texture descriptor for each image patch. The image patch size in this work is fixed at 25$\times$25 which is a trade-off between computational efficiency and description power. Empirical evaluation of the image patch size is conducted for the size range of 15 to 35 pixels using a small sub-sampled dataset for classification, as described later. Stable performance statistics are observed and quantitative experimental results using the default patch size of 25$\times$25 pixels are reported.

%%%%%%%%
%%%%%%%%%%%

\begin{figure}[t]
\begin{center}
\includegraphics[width=1.00\linewidth]{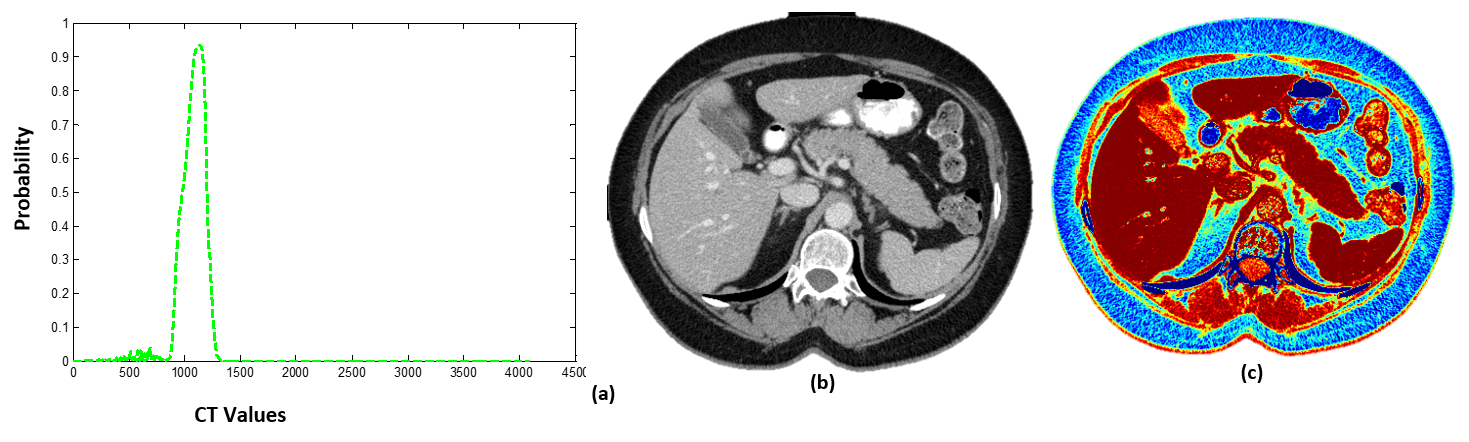}
\end{center}
\caption{Examples of computed KDE histogram (a) and sample 2D probability response map (c) of the original CT in (b). Red presents higher value; blue for lower probability.}\label{fig:KDE}
\end{figure}

2), A second feature group using the voxel intensity histograms of the ground-truth pancreas and the surrounding CT scans is built in the class-conditional probability density function (PDF) space. A kernel density estimator (KDE\footnote{\url{http://www.ics.uci.edu/~ihler/code/kde.html}}) is created using the voxel intensities from a subset of randomly selected patient CT scans. The KDE represents the CT intensity distributions of the positive $\left\{X^{+}\right\}$ and negative class $\left\{X^{-}\right\}$ of pancreas and non-pancreas voxels’ CT image information. \textcolor{LightRubineRed} {All voxels containing pancreas information are considered in the positive sample set, yet, since negative voxels far outnumber the positive only $5\%$ of the total number from each CT scan (by random resampling) is considered.} Let,  $\left\{X^{+}\right\} = \left(h_1^+,h_2^+,\cdots,h_n^+\right)$ and $\left\{X^{-}\right\} = \left(h_1^-,h_2^-,\cdots,h_m^-\right)$ where $h_n^+$  and $h_m^-$ represent the intensity values for the positive and negative pixel samples for all 26 patient CT scans over the entire abdominal CT Hounsfield range. The kernel density estimators $f^+ (X^+)=\frac{1}{n}\sum^{n}_{i=1}K\left(X^{+}-X^{+}_{i}\right)$
and $f^-(X^-)=\frac{1}{m}\sum^{m}_{j=1}K\left(X^{-}-X^{-}_{j}\right)$ are computed where $K(∙)$ is assumed to be \textcolor{LightRubineRed} {a Gaussian kernel with optimal computed bandwidth, for this data, of 3.039. Kernel sizes or bandwidth may be selected automatically using 1D Likelihood-based search, as provided by the used KDE toolkit.} The normalized likelihood ratio is calculated which becomes a probability value as a function of intensity in the range of $H=[0:1:4095]$. Thus, the probability of being considered pancreas is formulated as: $y^+=\frac{(f^+(X^+))}{(f^+(X^+)+f^-(X^-))}$. This function is converted as a pre-computed look-up table over $H=[0:1:4095]$, which allows very efficient $O(1)$ access time. Figure \ref{fig:KDE} depicts the normalized KDE histogram computed and sample 2D probability response result on a CT slice. High probability regions are red in color and low probabilities in blue. In the original CT intensity domain the mean, median and standard deviation (std) statistics over the full 25$\times$25 pixel range per image patch, \emph{P}, are extracted.

%%%%%%%%
%%%%%%%%%%%

3), Utilizing first the KDE probability response maps above and the superpixel CT masks described in Sec.~\ref{subsec:SP}, as underlying supporting masks to each image patch, the same KDE response statistics within the intersected sub-regions, P' of P, are extracted. The idea is that an image patch, P, may be divided into more than one superpixel. This set of statistics is calculated with respect to the most representative superpixel (that covers the patch center pixel). In this manner, object boundary-preserving intensity features are obtained \cite{kim}.

4), The final two features for each axial slice (in the patient volumes) are the normalized relative x-axis and y-axis positions $\epsilon[0,1]$, computed at each image patch center against the segmented body region (self-normalized\footnote{The axial reconstruction CT scans in our study have largely varying ranges or extends in the z-axis. If some anatomical landmarks, such as the bottom plane of liver, the center of kidneys, can be provided automatically, the anatomically normalized z-coordinate positions for superpixels can be computed and used as an additional spatial feature for RF classification.} to patients with different body masses to some extent). Once all of the features are concatenated together, a total of $46$ image patch-level features per superpixel are used to train a random forest (RF) classifier $C_p$. Image patch labels are obtained by directly borrowing the class information of their patch center pixels, based on the manual segmentation.

To resolve the data unbalancing issue between the positive (pancreas) and negative (non-pancreas) class patches (during RF training), the sample weights for two classes are normalized so that the sum of the sample weights for each class will reach the same constant (i.e., assuming a uniform prior). This means, when calculating the empirical two class probabilities at any RF leaf node, the positive and negative samples (reached at that node) contribute with different weights. Pancreas class patch samples are weighted much higher than non-pancreas background class instances since they are more rare. Similar sample reweighting scheme is also exploited in the regression forest for anatomy localization \cite{Criminisi}.

\begin{figure*}[t]
\centerline{
\includegraphics[width=1.50\columnwidth]{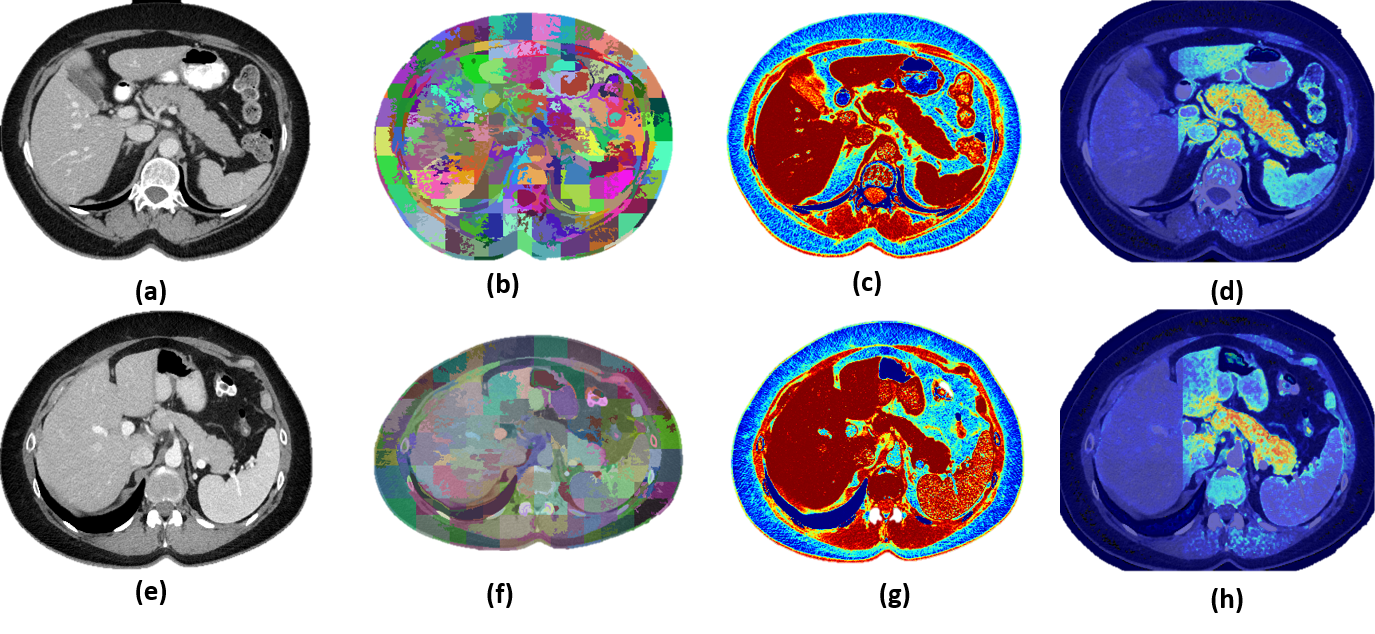}}
\caption{Two sample slices from different patients are shown in (a) and (e). The corresponding superpixels segmentation (b,f), KDE probability response maps (c, g) and RF patch-level probability response maps (d, h) are shown. In (c,g) and (d,h), red represents highest probabilities. In (d,h) the purple color represents areas where probabilities are so small and can be deemed insignificant areas of interest.}\label{fig:Overall}
\end{figure*}

Six-fold cross-validation for RF training is carried-out. Response maps are computed for the image patch-level classification and dense labeling. Fig. \ref{fig:Overall} (d) and (h) show sample illustrative slices from different patients. High probability corresponding to the pancreas is represented by the red color regions (the background is blue). The response maps (denoted as $P^{RF}$) allow several observations to be made. The most interesting is that the relative x and y positions as features allow for clearer spatial separation of positive and negative regions, via internal RF feature thresholding tests on them. The trained RF classifier is able to recognize the negative class patches residing in the background, such as liver, vertebrae and muscle using spatial location cues. In Fig. \ref{fig:Overall}(d,h) implicit vertical and horizontal decision boundary lines can be seen in comparison to Fig. \ref{fig:Overall}(c,g). This demonstrates the superior descriptive and discriminative power of the feature descriptor on image patches (P and P') than single pixel intensities. Organs with similar CT values are significantly depressed in the patch-level response maps.

%%%%%%%%
%%%%%%%%%%%

In summary, SIFT and its variations, e.g., D-SIFT have shown to be informative, especially through spatial pooling or packing \cite{Gilinsky}. A wide range of pixel-level correlations and visual information per image patch is also captured by the rest of 14 defined features. Both good classification specificity and recall have been obtained in cross-validation using Random Forest implementation of 50 trees and the minimum leaf size set as 150 (i.e., using the $treebagger(\bullet)$ function in Matlab).

\subsection{Patch-level Labeling via Deep Convolutional Neural Network: $P^{CNN}$} \label{subsec:CNN}

\begin{figure*}[t]%[htb!]
\centerline{	\includegraphics[width=1.50\columnwidth]{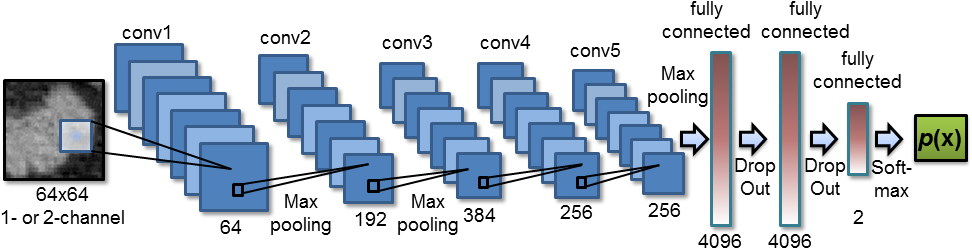}}
	\caption{\small The proposed CNN model architecture is composed of five convolutional layers with max-pooling and two fully-connected layers with DropOut \cite{srivastava2014dropout} connections. A final 2-way softmax layer gives a probability $p(x)$ of `pancreas' and `non-pancreas' per data sample (or image patch). The number and model parameters of convolutional filters and neural network connections for each layer are as shown.}
	\label{fig:convnet}
\end{figure*}

In this work, we use Convolutional Neural Network (CNN, or ConvNet) with a standard architecture for binary image patch classification. Five layers of convolutional filters first compute, aggregate and assemble the low level image features to more complex ones, in a layer-by-layer fashion. Other CNN layers perform max-pooling operations or consist of fully-connected neural network layers. The CNN model we adopted ends with a final two-way softmax classification layer for 'pancreas' and 'non-pancreas' classes (refer to Fig. \ref{fig:convnet}). The fully connected layers are constrained using ``DropOut'' in order to avoid over-fitting in training where each neuron or node has a probability of 0.5 to be reset with a 0-valued activation. DropOut is a method that behaves as a co-adaption regularizer when training the CNN \cite{srivastava2014dropout}. In testing, no DropOut operation is needed. Modern GPU acceleration allows efficient training and run-time testing of the deep CNN models. We use the publicly available code base of \textit{cuda-convnet2}\footnote{\url{https://code.google.com/p/cuda-convnet2}}.

\begin{figure*}[t]%[htb!]
\centerline{\includegraphics[width=1.50\columnwidth]{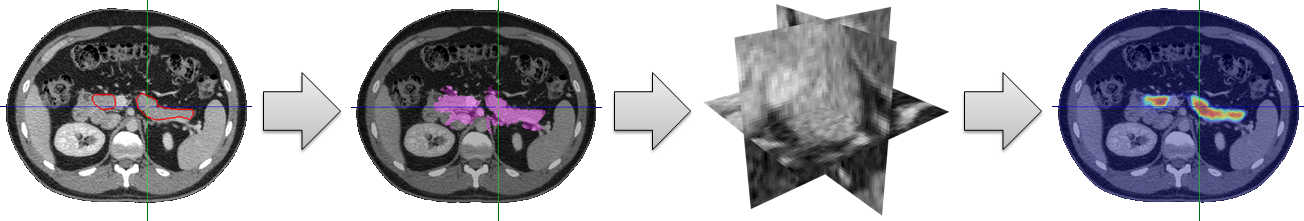}}
	\caption{\small Axial CT slice of a manual (gold standard) segmentation of the pancreas. From {\em Left} to {\em Right}, there are the ground-truth segmentation  contours (in red); RF based coarse segmentation $\{S_\mathrm{RF}\}$; a 2.5D input image patch to CNN and the deep patch labeling result using CNN. }%The shape and size of the pancreas can vary drastically between patients. CT densities within the pancreas can vary and the contrast to surrounding tissues can be low in CT.}					
	\label{fig:patch-convnet}
\end{figure*}

To extract dense image patch response maps, we use a straight-forward sliding window approach that extracts 2.5D image patches composed of axial, coronal and sagittal planes at any image positions (see Fig. \ref{fig:patch-convnet}). Deep CNN architecture can encode large scale image patches (even the whole
224$\times$224 pixel images \cite{krizhevsky2012imagenet}) very efficiently and no hard crafted image features are required any more. In this paper, the dimension of image patches for training CNN is 64$\times$64 pixels which is significantly larger than 25$\times$25 in Sec. \ref{subsec:patch}. The larger spatial scale or context is generally expected to achieve more accurate patch labeling quality. For efficiency reasons, we extract patches every $\ell$ voxels for CNN feedforward evaluation and then apply nearest neighbor interpolation\footnote{In our empirical testing, simple nearest neighbor interpolation seems sufficient due to the high quality of deep CNN probability predictions.} to estimate the values at skipped voxels. Three examples of dense CNN based image patch labeling are demonstrated in Fig. \ref{fig:patch-cnn-examples}. We denote the CNN model generated probability maps as $P^{CNN}$.

The computational expense of deep CNN patch labeling per patch (in a sliding window manner) is still higher than Sec. \ref{subsec:patch}. In practice, dense patch labeling by $P^{RF}$ runs exhaustively at 3 pixel interval but $P^{CNN}$ are only evaluated at pixel locations that pass the first stage of a cascaded random forest superpixel classification framework. \textcolor{LightRubineRed} {This process is detailed in Sec. \ref{subsec:Seg} where $C_{SP}^1$ is operated at a high recall (close to $100\%$) and low specificity mode to minimize the false negative rate (FNR) as the initial layer of cascade. The other important reason for doing so is to largely alleviate the training unbalance issue for $P^{CNN}$ in $C_{SP}^3$. After this initial pruning, the number ratio of non-pancreas versus pancreas superpixels changes from $>100$ to $\sim5$. The similar treatment is employed in our recent work \cite{Roth2015} where all  ``Regional CNN'' (R-CNN) based algorithmic variations \cite{Girshick14} for pancreas segmentation is performed after a superpixel cascading.} %A final 3D aggregation step can be applied as described in Sec. \ref{sec:3D_aggregation}.

\subsection{Superpixel-level Feature Extraction, Cascaded Classification and Pancreas Segmentation} \label{subsec:Seg}

%%%%%%%%
\begin{figure}[t]
\begin{center}
\includegraphics[width=0.48\linewidth]{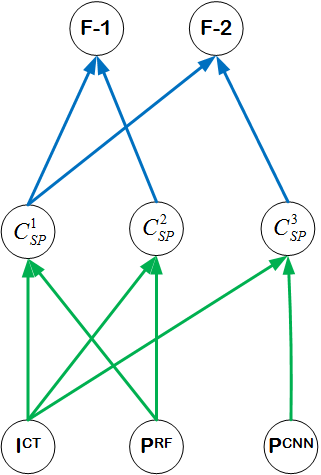}
\end{center}
\caption{The flow chart of input channels and component classifiers to form the overall frameworks 1 (F-1) and 2 (F-2). $I^{CT}$ indicates the original CT image channel; $P^{RF}$ represents the probability response map by RF based patch labeling in Sec. \ref{subsec:patch} and $P^{CNN}$ from deep CNN patch classification in Sec. \ref{subsec:CNN}, respectively. Superpixel level random forest classifier $C_{SP}^1$ is trained with all positive and negative superpixels in $I^{CT}$ and $P^{RF}$ channels; $C_{SP}^2$ and $C_{SP}^3$ are learned using only ``hard negatives'' and all positives, in the $I^{CT} \bigcup P^{RF}$ or $I^{CT} \bigcup P^{CNN}$ channels, respectively. Forming $C_{SP}^1 \mapsto C_{SP}^2$, or $C_{SP}^1 \mapsto C_{SP}^3$ into two overall cascaded models results in frameworks F-1 and F-2. \textcolor{LightRubineRed} {Note that F-1 and F-2 share the first layer of classification cascades to coarsely prune about $96\%$ initial superpixels using both intensity and $P^{RF}$ features (refer to Fig. \ref{fig:ROCPatch} {\bf Left}). Only intensity based statistical features for $C_{SP}^1$ produce significantly inferior results. On the other hand, $C_{SP}^3$ can be learned using all three available information channels of $I^{CT} \bigcup P^{RF} \bigcup P^{CNN}$ that will result in 36 superpixel-level features. Based on our initial empirical evaluation, $I^{CT} \bigcup P^{CNN}$ is as sufficient as using all three channels. This means that $I^{CT} \bigcup P^{CNN}$ seems the optimal feature channel combination or configuration considering both the classification effectiveness and model complexity. The coupling of $P^{CNN}$ into $C_{SP}^3$ consistently shows better segmentation results than $P^{RF}$ for $C_{SP}^2$ whereas $P^{CNN}$ is not powerful enough to be used alone.} } \label{fig:Classifier_Flow}
\end{figure}
%%%%%%%%%%%

In this section, we trained three different superpixel-level random forest classifiers of $C_{SP}~1$, $C_{SP}~2$ and $C_{SP}~3$. These three classifier components further formed two cascaded RF classification frameworks (F-1, F-2), as shown in Fig. \ref{fig:Classifier_Flow}. The superpixel labels are inferred from the overlapping ratio $r$ (defined in Sec. \ref{subsec:SP}) between the superpixel label map and the ground-truth pancreas mask. If $r\geq0.5$, the superpixel is positive while if $r\leq0.2$, the superpixel is assigned as negative. For the rest of superpixels that fall within $0.2<r<0.5$ (a relatively very small portion/subset of all superpixels), they are considered ambiguous and not assigned a label and as such not used in training.

Training $C_{SP}^1$ utilizes both the original CT image slices ($I^{CT}$ in Fig. \ref{fig:Classifier_Flow}) and the probability response maps ($P^{RF}$) via the hand-crafted feature based patch-level classification (i.e. Sec.\ref{subsec:patch}). The 2D superpixel supporting maps (i.e., Sec.\ref{subsec:SP}) are used for feature pooling and extraction on a superpixel level. The CT pixel intensity/attenuation numbers and the per-pixel pancreas class probability response values (from dense patch labeling of $P^{PF}$ or $P^{CNN}$ later) within each superpixel are treated as two empirical unordered distributions. Thus our superpixel classification problem is converted as modeling the difference between empirical distributions of positive and negative classes. We compute 1) simple statistical features of the 1st-4th order statistics such as mean, std, skewness, kurtosis \cite{Groeneveld} and 2) histogram-type features of eight percentiles $\left(20\%, 30\%,\ldots,90\%\right)$, per distribution in intensity or $P^{RF}$ channel, respectively. Once concatenated, the resulted 24 features for each superpixel instance is fed to train random forest classifiers.

Due to the highly unbalanced quantities between foreground (pancreas) superpixels and background (the rest of CT volume) superpixels, a two-tiered cascade of random forests are exploited to address this type of rare event detection problem \cite{Viola}. In a cascaded classification, $C_{SP}^1$ once trained is applied exhaustively on scanning all superpixels in an input CT volume. Based on the receiver operating characteristic (ROC) curves in Fig. \ref{fig:ROCPatch} (Left) for $C_{SP}^1$, we can safely reject or prune $97\%$ negative superpixels while maintaining nearly $\sim100\%$ recall or sensitivity. The remained $3\%$ negatives, often referred as ``hard negatives'' \cite{Viola}, along with all positives are employed to train the second $C_{SP}^2$ in the same feature space. Combining $C_{SP}^1$ and $C_{SP}^2$ is referred to as Framework 1 (F-1) in the subsequent sections.

%%%%%%%%%%%

Similarly, we can train a random forest classifier $C_{SP}^3$ by replacing $C_{SP}^2$'s feature extraction dependency on the $P^{RF}$ probability response maps, with the deep CNN patch classification maps of $P^{CNN}$. The same 24 statistical moments and percentile features per superpixel, from two information channels $I^{CT}$ and $P^{CNN}$, are extracted to train $C_{SP}^3$. Note that the CNN model that produces $P^{CNN}$ is trained with the image patches sampled from only ``hard negative'' and positive superpixels (aligned with the second-tier RF classifiers $C_{SP}^2$ and $C_{SP}^3$). For simplicity, $P^{RF}$ is only trained once with all positive and negative image patches. This will be referred to as Framework 2 (F-2) in the subsequent sections. F-1 only use $P^{RF}$ whereas F-2 depends on both $P^{RF}$ and $P^{CNN}$ (with a little extra computational cost).

The flow chart of frameworks 1 (F-1) and 2 (F-2) is illustrated in Fig. \ref{fig:Classifier_Flow}. The two-level cascaded random forest classification hierarchy is found empirically to be sufficient (although a deeper cascade is possible) and implemented to obtain F-1: $C_{SP}^1$ and $C_{SP}^2$, or F-2: $C_{SP}^1$ and $C_{SP}^3$. The binary 3D pancreas volumetric mask is obtained by stacking the binary superpixel labeling outcomes (after $C_{SP}^2$ in F-1 or $C_{SP}^3$ in F-2) for each 2D axial slice, followed by 3D connected component analysis implemented in the end. By assuming the overall pancreas connectivity of its 3D shape, the largest 3D connected component is kept as the final segmentation. The binarization thresholds of random forest classifiers in $C_{SP}^2$ and $C_{SP}^3$ are calibrated using data in the training folds in 6-fold cross-validation, via a simple grid search. \textcolor{LightRubineRed} {$C_{SP}^3$ can be learned using all three available information channels of $I^{CT} \bigcup P^{RF} \bigcup P^{CNN}$ that will produce 36 superpixel-level features. Based on our initial empirical evaluation, $I^{CT} \bigcup P^{CNN}$ is as sufficient as using all three channels. As demonstrated in Sec. \ref{sec:Results2}, $C_{SP}^3$ outperforms $C_{SP}^2$ mainly due to the sake of $P^{CNN}$ instead of $P^{RF}$. However $P^{CNN}$ is not powerful enough to be used alone in $C_{SP}^3$. In \cite{Roth2015}, standalone $Patch$-ConvNet dense probability maps (without any post-processing) are processed for pancreas segmentation after using (F-1) as an initial cascade. The corresponding pancreas segmentation performance is not as accuracy as (F-1) or (F-2). This finding is in analogy to \cite{van2015off,barchest2015} where hand-crafted features need to be combined with deep image features to improve pulmonary nodule detection in chest CT scans or chest pathology detection using X-rays. Refer to Sec. \ref{sec:Results2} for detailed comparison and numerical results. Recent computer vision work also demonstrate the performance improvement when combining hand-crafted and deep image features for image segmentation \cite{Paisitkriangkrai2015} and video action recognition \cite{Lan2015} tasks. }

\section{DATA AND EXPERIMENTAL RESULTS}\label{sec:Results}
\subsection{Imaging Data}
80 3D abdominal portal-venous contrast enhanced CT scans ($\sim70$ seconds after intravenous contrast injection) acquired from 53 male and 27 female subjects are used in our study for evaluation. 17 of the subjects are from a kidney donor transplant list of healthy patients that have abdominal CT scans prior to nephrectomy. The remaining 63 patients are randomly selected by a radiologist from the Picture Archiving and Communications System (PACS) on the population that has neither major abdominal pathologies nor pancreatic cancer lesions. The CT datasets are obtained from National Institutes of Health Clinical Center. Subjects range in the age from 18 to 76 years with a mean age of $46.8\pm16.7$. Scan resolution has 512$\times$512 pixels (varying pixel sizes) with slice thickness ranging from $1.5-2.5$ mm on Philips and Siemens MDCT scanners. The tube voltage is 120 kVp. Manual ground-truth segmentation masks of the pancreas for all 80 cases are provided by a medical student and verified/modified by a radiologist.

\subsection{Experiments}\label{sec:Results2}

Quantitative experimental results are assessed using six-fold cross-validation, as described in Sec. \ref{subsec:patch} and Sec. \ref{subsec:Seg}. Several metrics to evaluate the accuracy and robustness of the methods are computed. The Dice similarity index which interprets the overlap between two sample sets, $SI=2(|A \cap B|)/(|A|+|B|)$  where $A$ and $B$ refer to the algorithm output and manual ground-truth 3D pancreas segmentation, respectively. The Jaccard index (JI) is another statistic used to compute similarities between the segmentation result against the reference standard, $JI=(|A \cap B|)/(|A \cup B|)$, called ``intersection over union'' in the PASCAL VOC challenges \cite{Everingham2015,Carreira12a}. The volumetric recall (i.e. sensitivity) and precision values are also reported.

\begin{figure*}[t]
\centerline{
\begin{tabular}{cc}
\includegraphics[width=1.36\columnwidth]{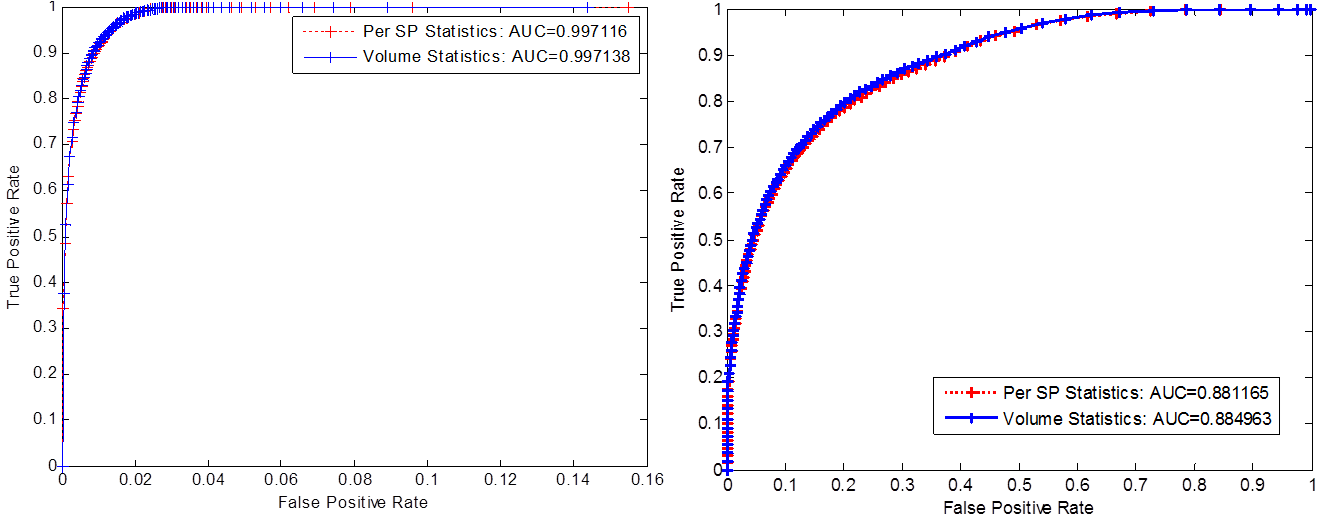} &
\includegraphics[width=0.68\columnwidth]{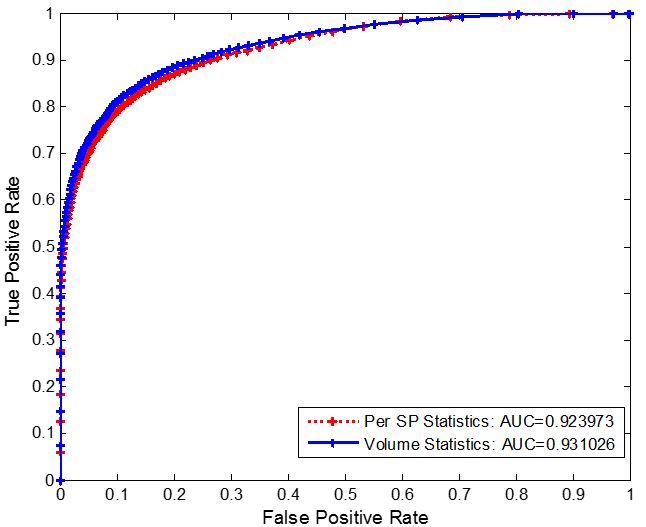} \\
\end{tabular}
}
\caption{ROC curves to analyze the superpixel classification results, in a two layer cascade of RF classifiers: ({\bf Left}) the first layer classifier, $C_{SP}^1$  and ({\bf Middle}) the second layer classifier, $C_{SP}^2$; ({\bf Right}) the alternative second layer classifier, $C_{SP}^3$. Red plots are using each superpixel as a count to calculate sensitivity and specificity. In blue plots, superpixels are weighted by their sizes (e.g., numbers of pixels and pixel sizes) for sensitivity and specificity calculation.}\label{fig:ROCPatch}
\end{figure*}

\begin{figure*}[ht]
\centerline{
\begin{tabular}{ccc}
    \includegraphics[width=0.46\columnwidth]{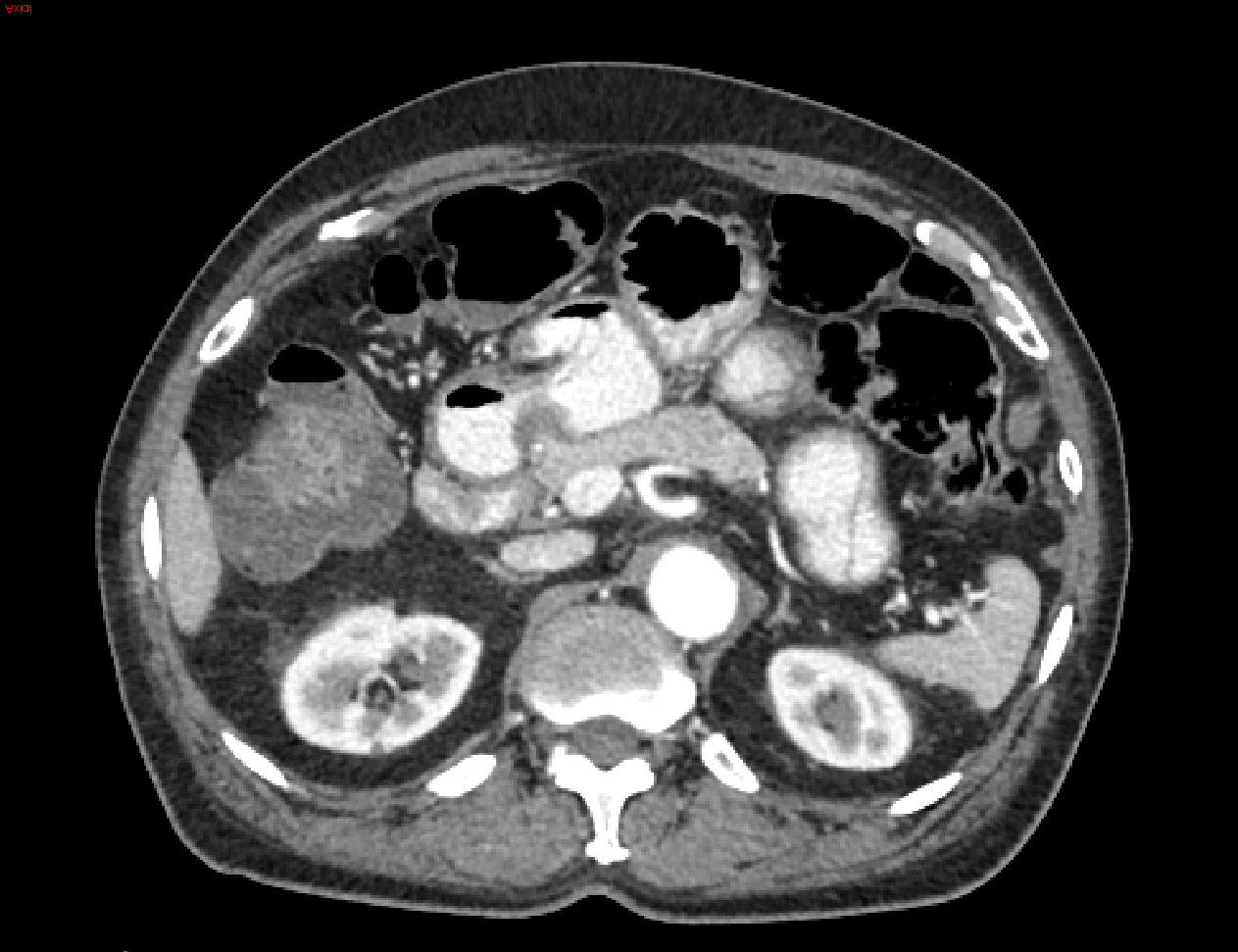} &
    \includegraphics[width=0.46\columnwidth]{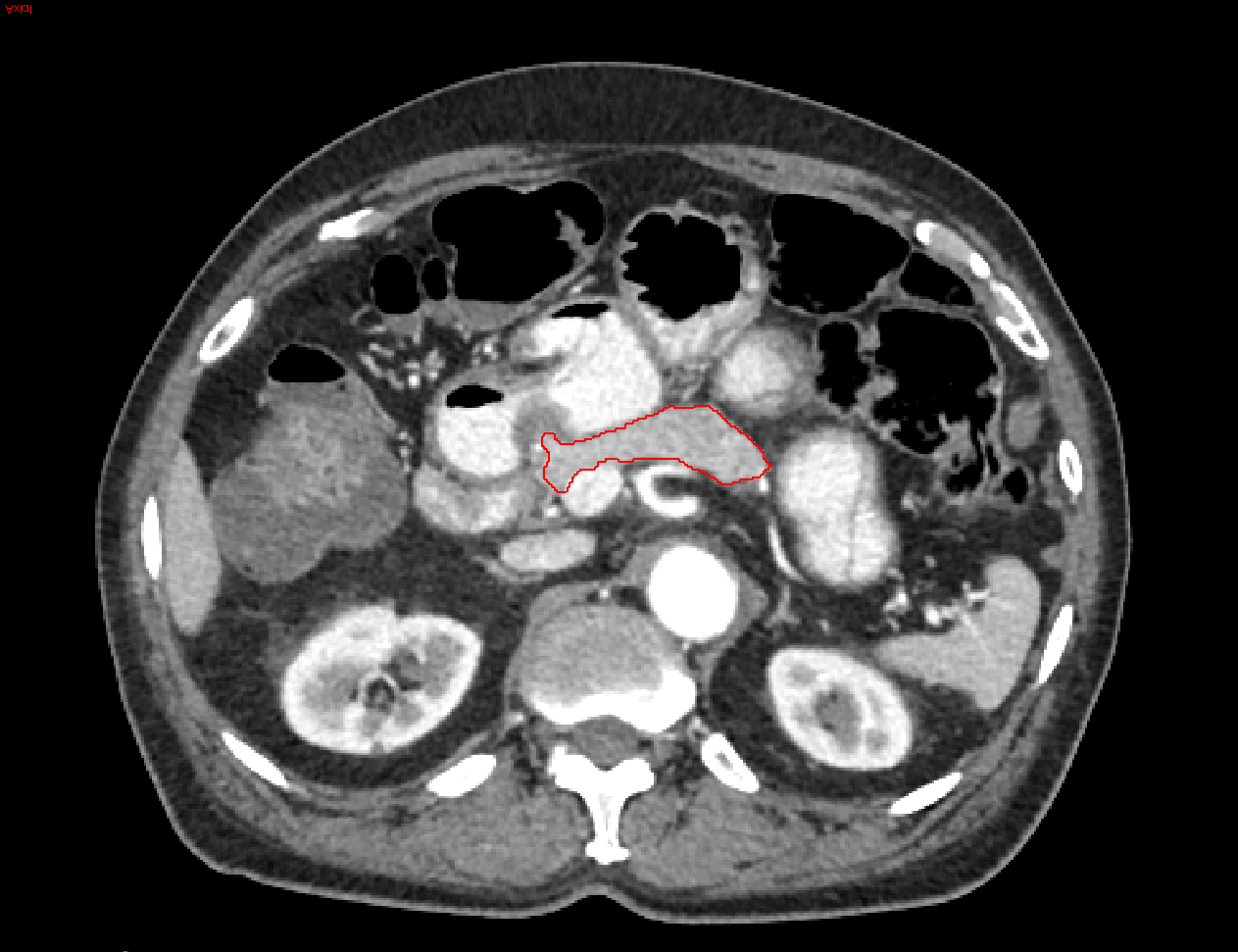} &
		\includegraphics[width=0.46\columnwidth]{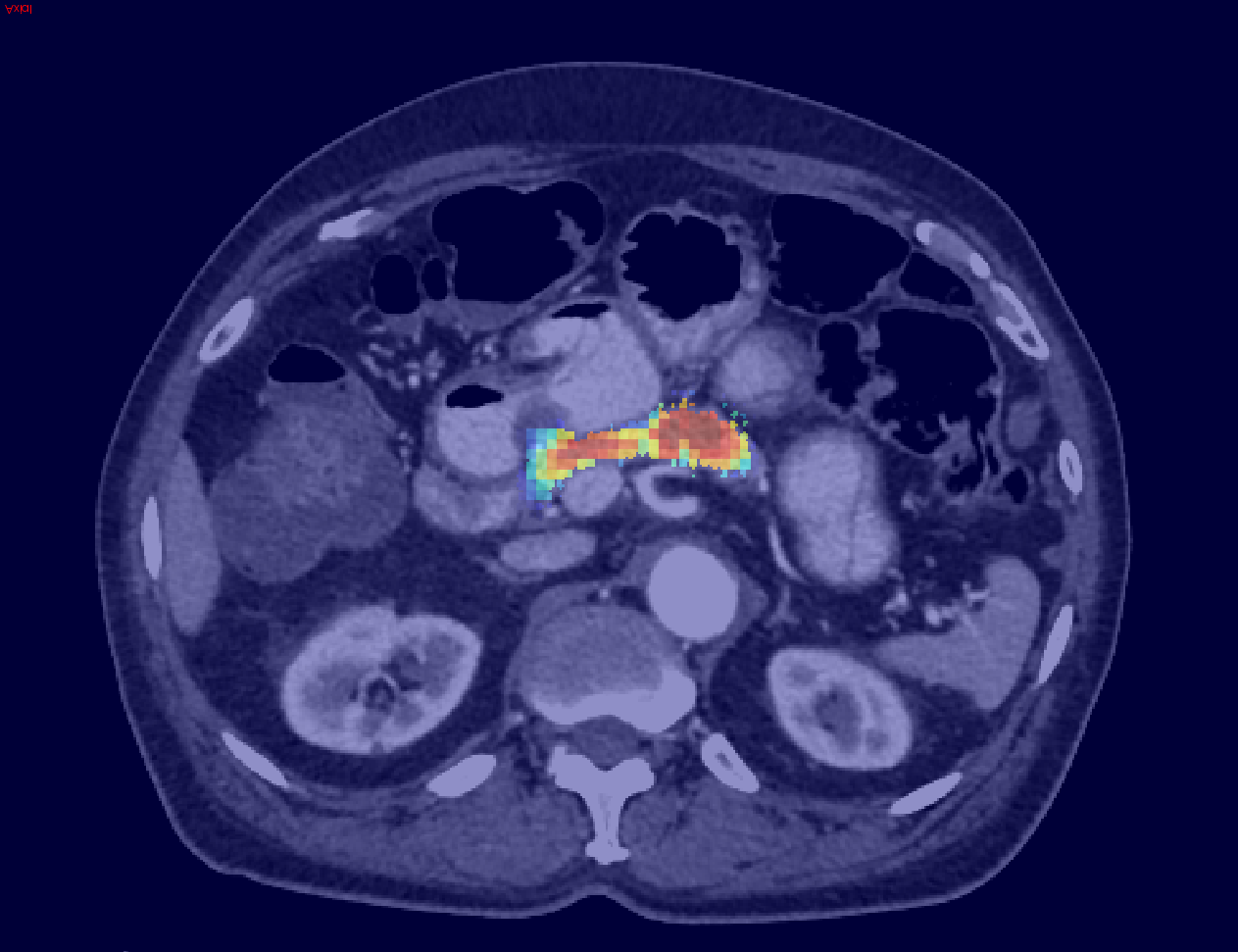} \\
		\includegraphics[width=0.46\columnwidth]{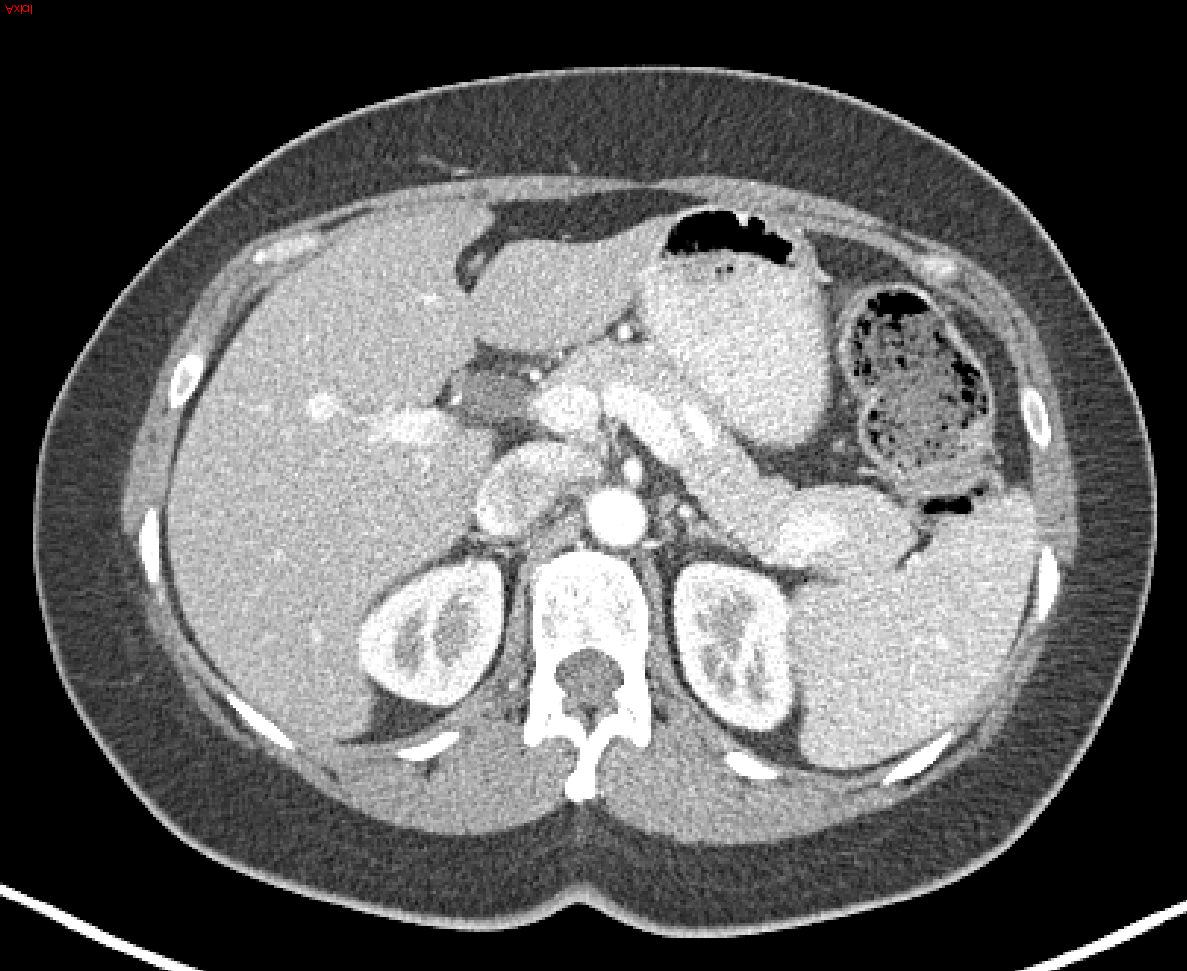} &
    \includegraphics[width=0.46\columnwidth]{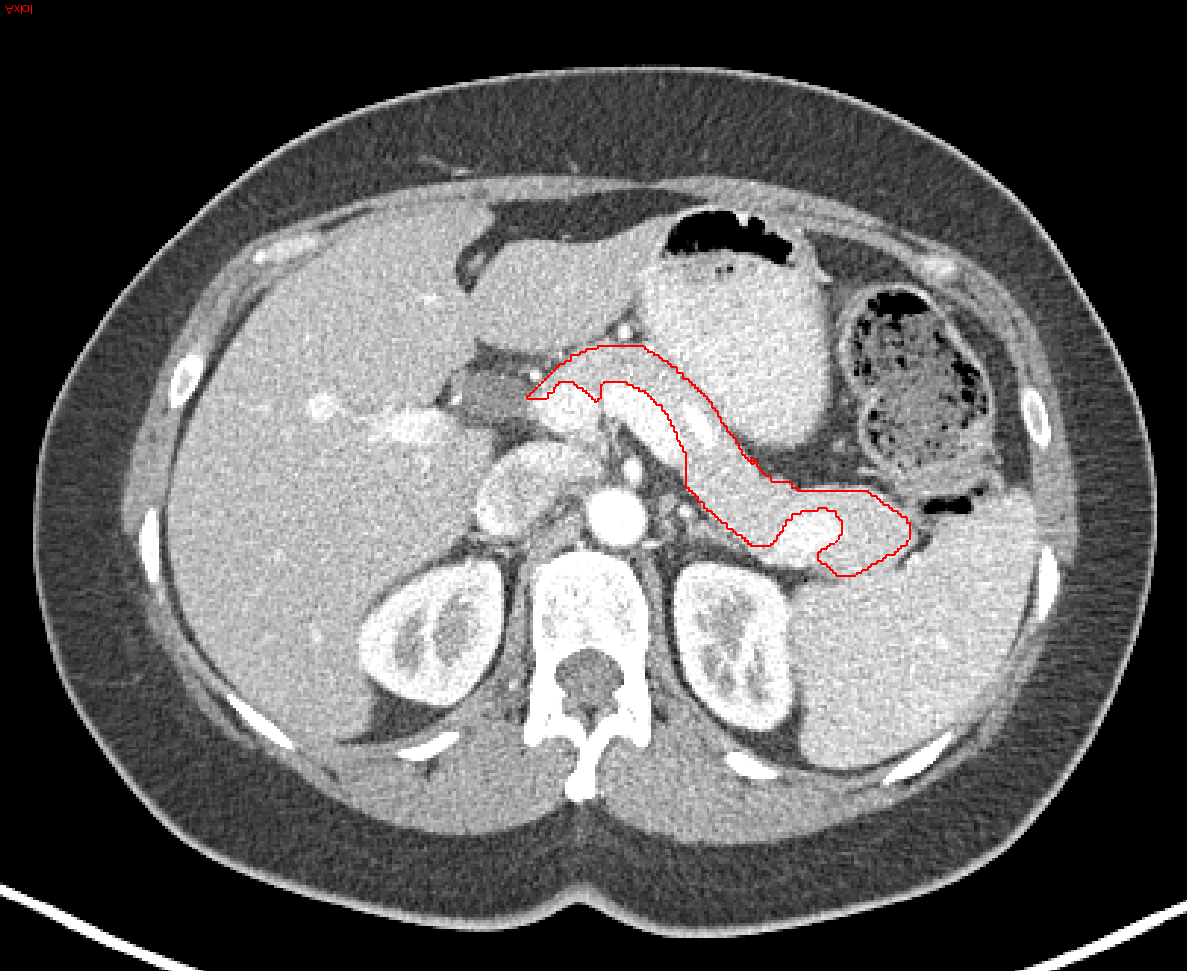} &
		\includegraphics[width=0.46\columnwidth]{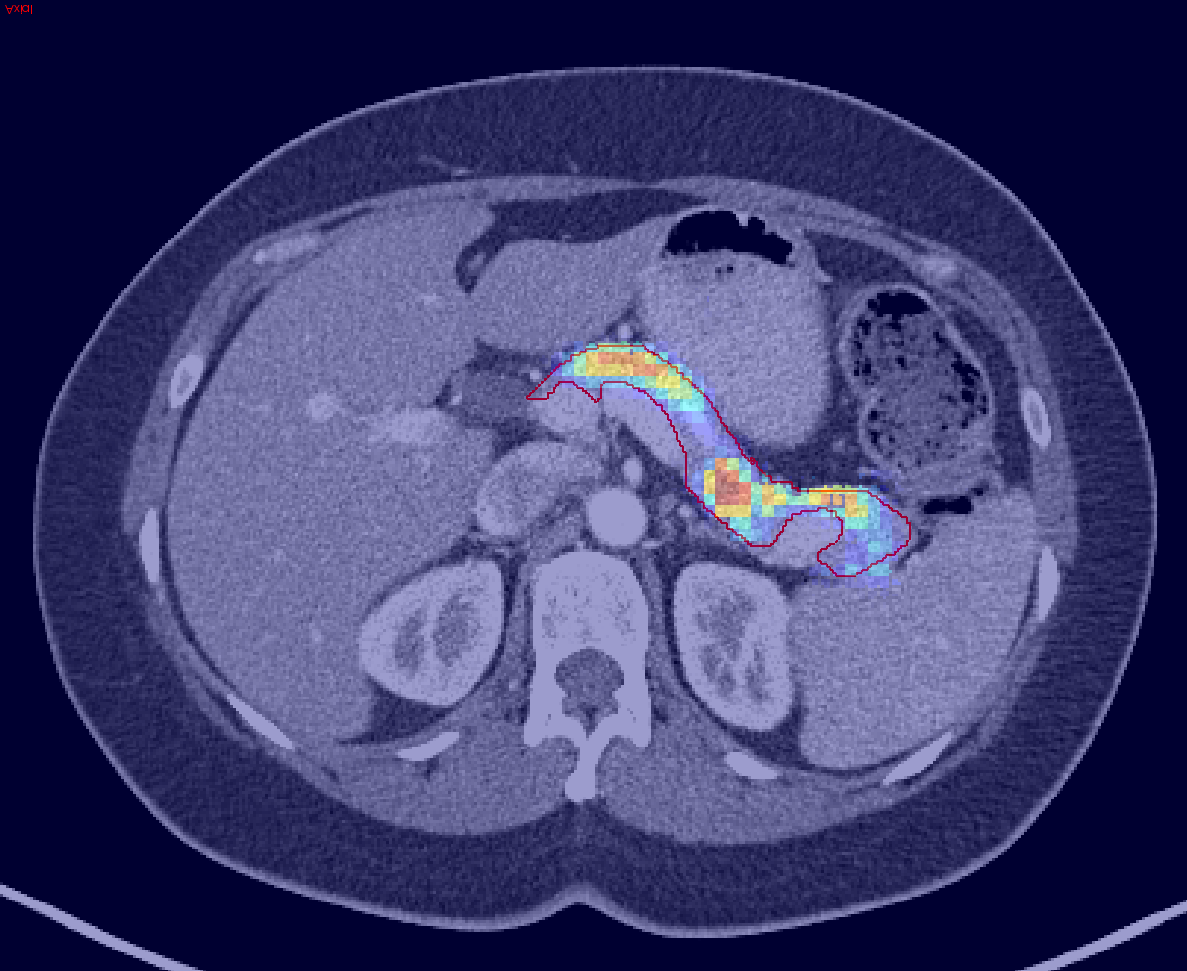} \\
		\includegraphics[width=0.46\columnwidth]{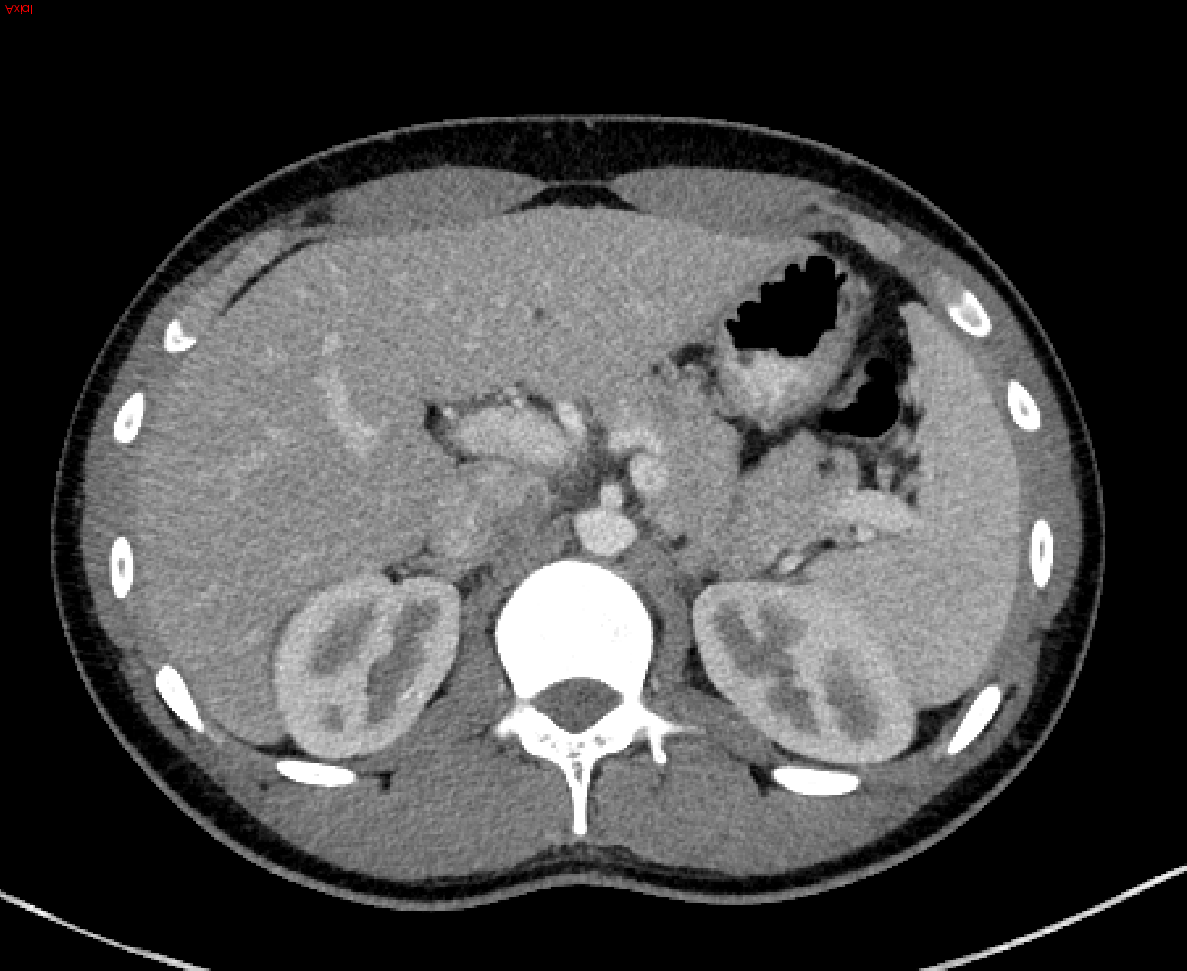} &
    \includegraphics[width=0.46\columnwidth]{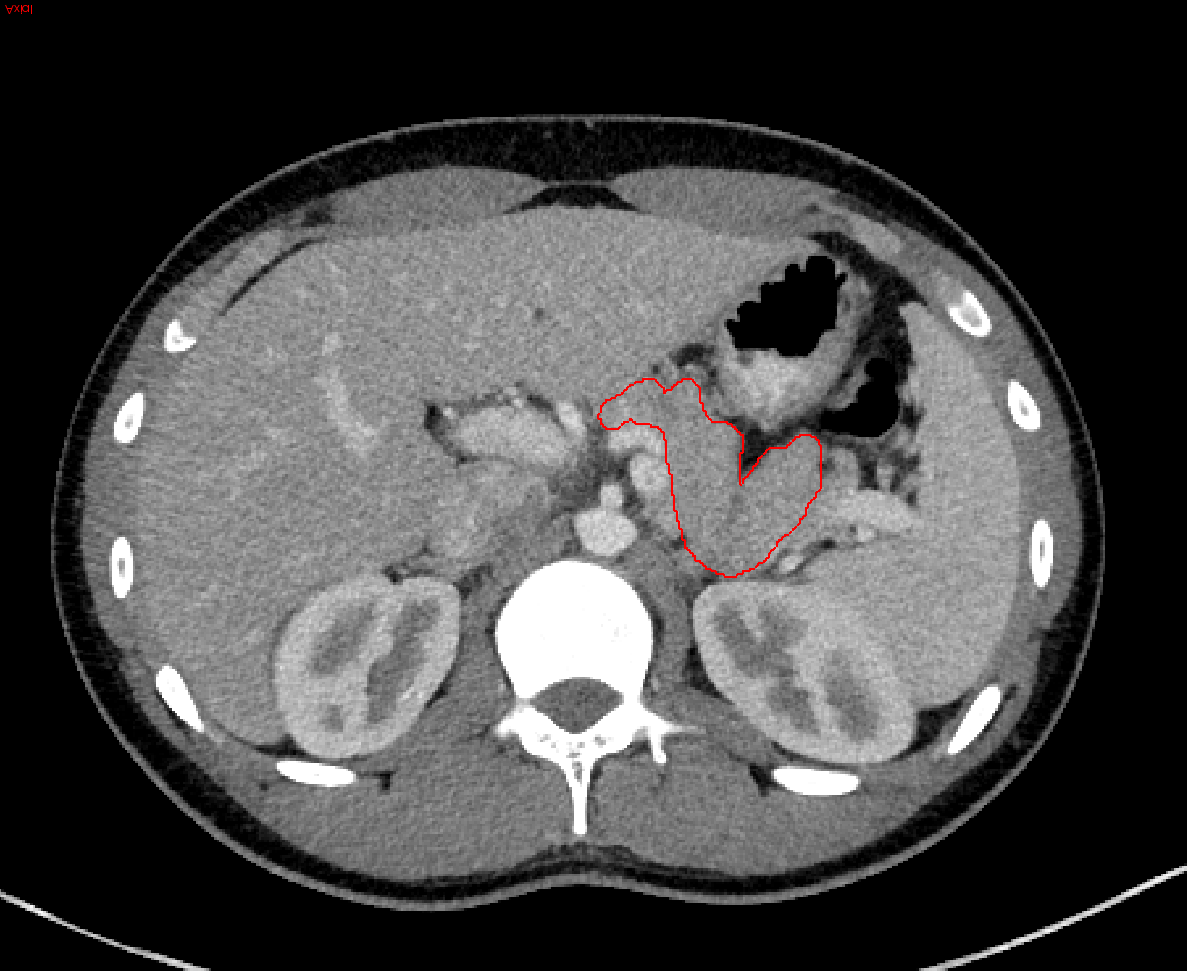} &
		\includegraphics[width=0.46\columnwidth]{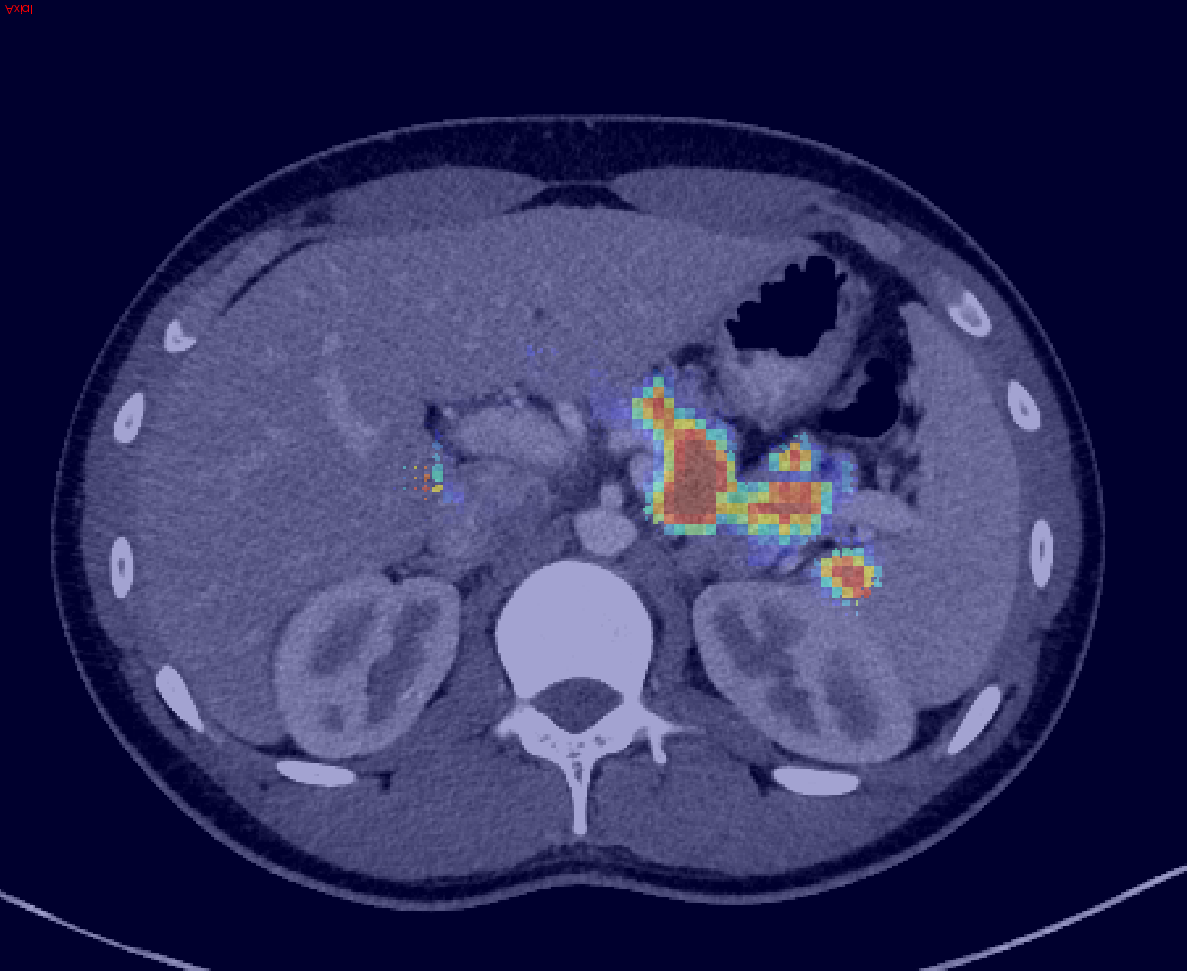} \\
    \end{tabular}
}
\caption{Three examples of deep CNN based image patch labeling probability response maps per row. Red color shows stronger pancreas class response and blue presents weaker response. From Left, Center to Right are the original CT image, CT image with annotated pancreas contour in red, and CNN response map overlaid CT image.} %(c) ROC curves of per-$V_i$ or volume $mm^3$ statistics when only $12$ boundary statistics features are used.}
\label{fig:patch-cnn-examples}
\end{figure*}

Fig. \ref{fig:ROCPatch} illustrates the ROC curves for 6-fold cross-validation of the two-tiered superpixel-level classifiers $C_{SP}^1$ and $C_{SP}^2$, $C_{SP}^3$ to assemble our frameworks F-1 and F-2, respectively. Red plots use each superpixel as a count to compute sensitivity and specificity. In blue plots, superpixels are weighted by their sizes (e.g., numbers of pixels and pixel sizes) for sensitivity and specificity calculation. The Area Under Curve (AUC) values of $C_{SP}^2$ are noticeably lower than $C_{SP}^1$ AUC values (0.884 versus 0.997), indicating which is much harder to train since it employs the ``hard negatives'' as negative samples what are classified positively by $C_{SP}^1$. Random Forest classifiers with $50\sim200$ trees are evaluated, with similar empirical performances. In $C_{SP}^3$, the dense patch-level image labeling (in the second level of cascade) is conducted by Deep Convolutional Neural Network (i.e., $Patch$-ConvNet) to generate $P^{CNN}$. Three examples of dense CNN based image patch labeling are demonstrated in Fig. \ref{fig:patch-cnn-examples}. \textcolor{LightRubineRed} {The AUC value of $C_{SP}^3$ by swapping the probability response maps from $P^{RF}$ to $P^{CNN}$ does improve to 0.931, compared to 0.884 using $C_{SP}^2$ in the pixel-weighted volume metric. This demonstrates the performance benefit of using CNN for dense patch labeling (Sec. \ref{subsec:CNN}) versus hand-crafted image features (Sec. \ref{subsec:patch}). See Fig. \ref{fig:ROCPatch} ({\bf Right}) and ({\bf Middle}), respectively. Standalone $Patch$-ConvNet dense probability maps can be smoothed and thresholded for pancreas segmentation as reported in \cite{Roth2015} where Dice coefficients $60.9 \pm 10.4\%$ are achieved. When there is only 12 features extracted from $P^{CNN}$ maps for $C_{SP}^3$, the final pancreas segmentation accuracy drops to  $64.5 \pm 12.3\%$ in Dice scores, compared to F-1 ($68.8  \pm 25.6\%$) and F-2 ($ 70.7 \pm 13.0\%$) in Table \ref{F1Eval}. Similarly, recent work \cite{van2015off,barchest2015} observe that deep CNN image features should be combined with hand-crafted features for better performances in computer-aided detection tasks.}

%%%%%%%%

\begin{figure*}[t]
\centerline{
\includegraphics[width=1.58\columnwidth]{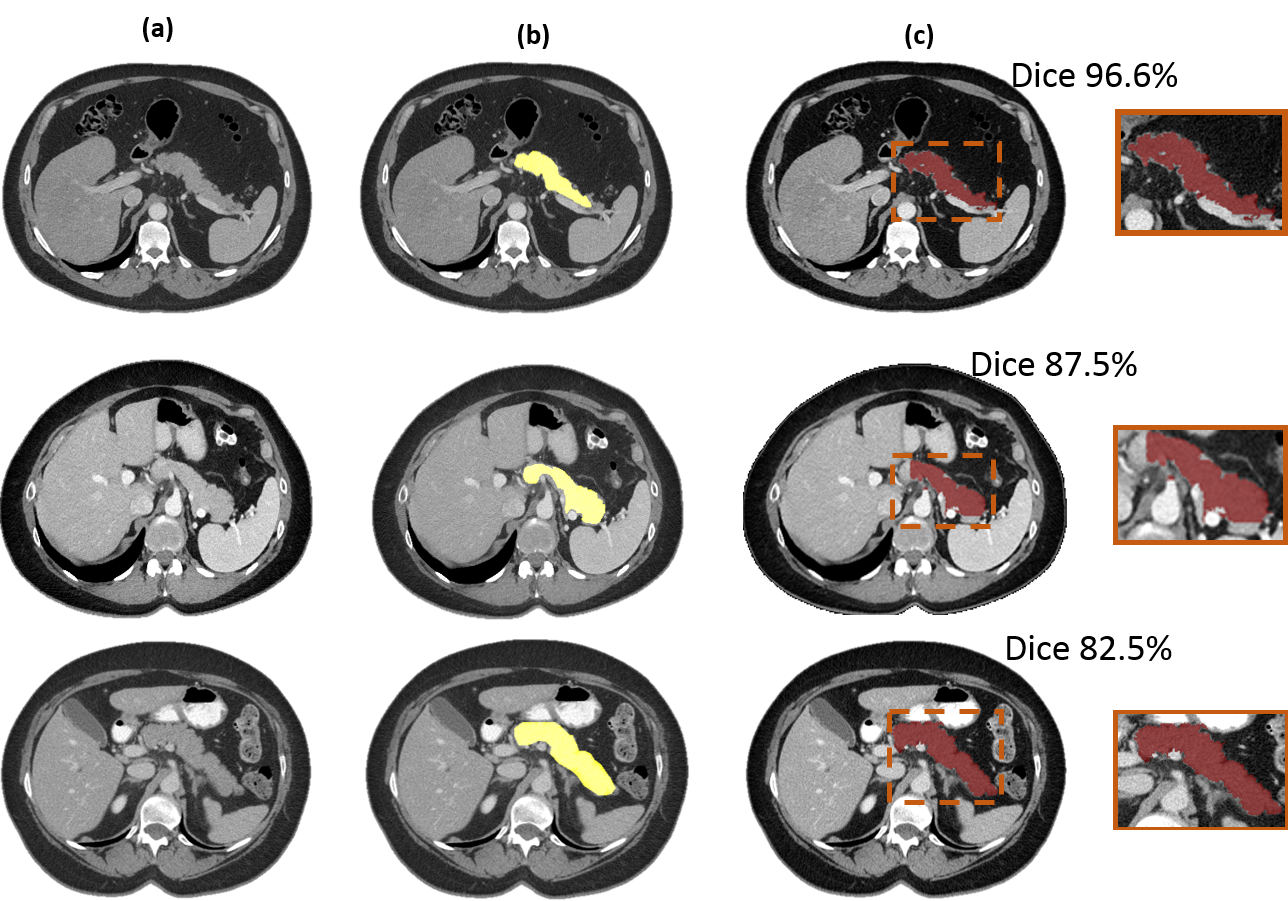}}
\caption{Pancreas segmentation results with the computed Dice coefficients for one good ({\bf Top Row}) and two fair ({\bf Middle, Bottom Rows}) segmentation examples. Sample original CT slices for both patients are shown in ({\bf Left Column}) and the corresponding ground truth manual segmentation in ({\bf Middle Column}) are in yellow. Final computed segmentation regions are shown in red in ({\bf Right Column}) with Dice coefficients for the volume above each slice. The zoomed-in areas of the slice segmentation in the orange boxes are shown to the right of the image.}\label{fig:ResF1}
\end{figure*}

Next, the pancreas segmentation performance evaluation is conducted in respect to the total number of patient scans used for the training and testing phases. Using our framework F1 on 40, 60 and 80 (i.e. $50\%$, $75\%$ and $100\%$ of the total 80 datasets) patient scans, the Dice, JI, Precision and Recall are computed under six-fold cross-validation. %The result numbers as well as the patient datasets used for training and validation changed for each case (i.e. training and validation datasets were not fixed as more data was added, the same randomization process occurred for the three intervals under the six-fold cross validation).
Table \ref{F1Eval} shows the computed results using image patch-level features and multi-level classification (i.e., performing $C_{SP}^1$ and $C_{SP}^2$ on $I^{CT}$ and $P^{RF}$) and how performance changes with the additions of more patients data. Steady improvements of $\sim4\%$ in the Dice coefficient and $\sim5\%$ for the Jaccard index are observed, from 40 to 60, and 60 to 80. Fig. \ref{fig:ResF1} illustrates some sample final pancreas segmentation results from the 80 patient execution (i.e. Test 3 in Table \ref{F1Eval}) for two different patients. The results are divided into three categories: good, fair and poor. The good category refers to the computed Dice coefficient above $90\%$ (of 15 patients), fair result as $50\% \leq Dice\geq 90\%$ (49 patients) and poor for Dice $<50\%$ (16 patients). %Fig. \ref{fig:ResF2} shows an example of pancreas segmentation result, comparing with the corresponding ground truth in 3D, with a surface-to-surface distance map.

%%%%%%%%
%%%%%%%%%%%

%%%%%%%%
%\begin{figure}[t]
%\centerline{
%\includegraphics[width=1.00\linewidth]{3DPancreasSegmentation.png}}
%\caption{Pancreas segmentation result comparison in 3D. The computed 3D pancreas segmentation mask (before 3D connected component based removal) is shown in ({\bf Left}) and the corresponding ground truth segmentation mask in ({\bf Middle}) are in red. Their surface-to-surface distance map overlaid on the ground truth mask is demonstrated in ({\bf Right}). The red color illustrates higher difference and green for smaller distance.}\label{fig:ResF2}
%\end{figure}
%%%%%%%%%%%

%%%%%%%%
\begin{figure*}[t]
\centerline{
\includegraphics[width=1.68\columnwidth]{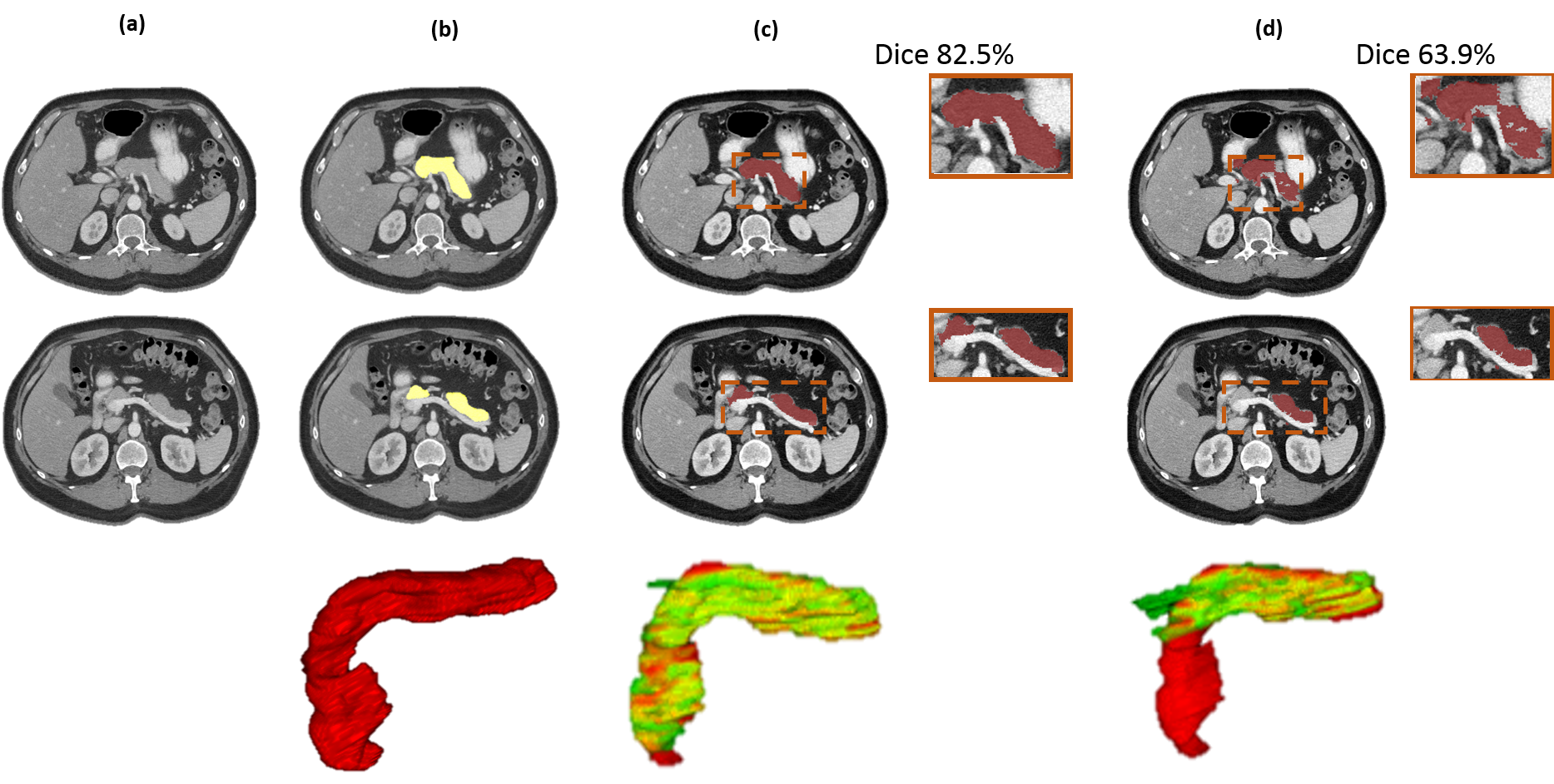}}
\caption{Examples of pancreas segmentation results using F-1 and F-2 with the computed Dice coefficients for one patient. Original CT slices for the patient are shown in {\bf Column (a)} and the corresponding ground truth manual segmentation in {\bf Column (b)} are in yellow. Final computed segmentation using F-2 and F-1 are shown in red in {\bf Columns (c,d)} with Dice coefficients for the volume above first slice. The zoomed-in areas of the slice segmentation in the orange boxes are shown to the right of the images. Their surface-to-surface distance map overlaid on the ground truth mask is demonstrated in {\bf Columns (c,d) Bottom} and the corresponding ground truth segmentation mask in {\bf Column (b) Bottom} are in red.  The red color illustrates higher difference and green for smaller distance.}\label{fig:ComparisonF1F2}
\end{figure*}
%%%%%%%%%%%

Then, we evaluate the difference of the proposed F-1 versus F-2 on 80 patients, using the same four metrics (i.e., Dice, JI, precision and recall). Table \ref{F1Eval} shows the comparison results. The same six-fold cross validation criterion is employed so that direct comparisons can be made. From the table, it can be seen that about $2\%$ increase in the Dice coefficient was obtained by using F-2, but the main improvement can be noticed in the minimum values (i.e., the lower performance bound) for each of the metrics. Usage of deep patch labeling prevents the case of no pancreas segmentation while keeping slightly higher mean precision and recall values. The standard deviations also dropped nearly $50\%$ comparing F-1 to F-2 (from 25.6\% to 13.0\% in Dice; and 25.4\% to 13.6\% in JI). Note that F-1 has the similar standard deviation ranges with the previous methods \cite{Wang,Wolz,Chu,Wolz12,Shimizu} and F-2 significantly improves upon all of them. From Fig. \ref{F1Eval} and Fig. \ref{fig:Overall} it can be inferred that using the relative x-axis and y-axis positions as features aided in reducing the overall false negative rates. Based on Table \ref{F1Eval}, we observe that F-2 provides consistent performance improvements over F-1, which implies that CNN based dense patch labeling shows more promising results (Sec. \ref{subsec:CNN}) than the conventional had-crafted image features and random forest patch classification alone (Sec. \ref{subsec:patch}). \textcolor{LightRubineRed} {Fig. \ref{fig:ComparisonF1F2} depicts an example patient where F-2 Dice score is improved by $18.6\%$ over F-1 (from $63.9\%$ to $82.5\%$). In this particular case, the close proximity of the stomach and duodenum to the pancreas head in particular proves challenging for F-1 without the CNN counterpart to distinguish. The surface-to-surface overlays illustrates how both frameworks compare to the ground truth manual segmentation.}

%Then, we evaluate the difference of the proposed framework 1 (F-1) versus framework 2 (F-2) on 80 patients, using the same four metrics (i.e., Dice, JI, precision and recall). Table \ref{F1Eval} shows the comparison results. The same six-fold cross validation criterion is employed so that direct comparisons can be made. From the table, it can be seen that about $2\%$ increase in the Dice coefficient was obtained by using Framework 2, but the main improvement can be noticed in the minimum values (i.e., the lower performance bound) for each of the metrics. Usage of deep patch labeling prevents the case of no pancreas segmentation while keeping slightly higher mean precision and recall values. The standard deviations also dropped nearly $50\%$ comparing F-1 to F-2 (from 25.6\% to 13.0\% in Dice; and 25.4\% to 13.6\% in JI). Note that F-1 has the similar standard deviation ranges with the previous methods \cite{Wang,Wolz,Chu,Wolz12,Shimizu}and F-2 significantly improves upon all of them. Based on Table \ref{F1Eval}, we observe that F-2 provides consistent performance improvements over F-1, which implies that CNN based dense patch labeling shows more promising results (Sec. \ref{subsec:CNN}) than the conventional had-crafted image features and random forest patch classification  alone (Sec. \ref{subsec:patch}). %Further examination into the sub-connectivity processes of the framework are needing further examination.

F-1 performs comparably to the state-of-the-art pancreas segmentation methods while F-2 slightly but consistently outperform others, even under six-fold cross-validation (CV) instead of the ``leave-one-patient-out'' (LOO) used in \cite{Wang,Wolz,Chu,Wolz12,Shimizu,Okada}. %The superpixel maps for each patient undergo a spatial support region reduction using its corresponding classification image patch-level probability response maps described in Sec.\ref{subsec:SP}.
\textcolor{LightRubineRed}{Note that our results are not directly or strictly comparable with \cite{Wang,Wolz,Chu,Wolz12,Shimizu,Okada} since different datasets are used for evaluation. If under the same six-fold cross-validation, our bottom-up segmentation method can significantly outperform an implemented version of ``multi-atlas and label fusion'' (MALF) based on \cite{Modat2010,Wang2012}, on the pancreas segmentation dataset studied in this paper. Details are provided later in this section.} Table \ref{Table:LitComp} reflects the comparison of Dice, JI, precision and recall results, between our methods of F-1, F-2 and other approaches, in multi-atlas registration and label fusion based multi-organ segmentation \cite{Wang,Wolz,Chu,Wolz12,Okada} and multi-phase single organ (i.e., pancreas) segmentation \cite{Shimizu}. Previous numerical results are found from the publications \cite{Wang,Wolz,Chu,Wolz12,Shimizu,Okada}. We choose the best result out of different parameter configurations in \cite{Chu}. Based on 80 CT datasets, our results are comparable and slightly better than the recent state-of-the art work \cite{Wang,Wolz,Chu,Wolz12,Shimizu}. For example, Dice coefficients of $68.8\%\pm 25.6\%$ using F-1 and $70.7\%\pm 13.0\%$ using F-2 are obtained (6-fold CV), versus $69.6\% \pm 16.7\%$ in \cite{Wolz}, $65.5\%$ in \cite{Wolz12}, $65.5\% \pm 18.6\%$ in \cite{Wang} and $69.1\% \pm 15.3\%$ in \cite{Chu} (LOO).

%%%%%%%%%%%%%%%%%%%%
\begin{table*}
\centerline{
  \begin{tabular}{cccccc}
  \hline
  ~& N &  SI (\%) &  JI (\%) &  Precision (\%) &  Recall (\%) \\ \hline
  F-1 & 40&
  $60.4  \pm 22.3 \left[2.0,~ 96.4\right]$&
  $46.7  \pm 22.8 \left[0,~  93.0\right]$&
  $55.6  \pm 29.8 \left[1.2,~ 100\right]$&
  $80.8  \pm 21.2 \left[4.8,~ 99.8\right]$\\ \hline
  F-1 & 60&
  $64.9  \pm 22.6 \left[0,~ 94.2\right]$&
  $51.7  \pm 22.6 \left[0,~ 89.1\right]$&
  $70.3  \pm 29.0 \left[0,~ 100\right]$&
  $69.1  \pm 25.7 \left[0,~ 98.9\right]$\\ \hline
  F-1 & 80&
  $68.8  \pm 25.6 \left[0,~ 96.6\right]$&
  $57.2  \pm 25.4 \left[0,~ 93.5\right]$&
  $71.5  \pm 30.0 \left[0,~ 100\right]$&
  $72.5  \pm 27.2 \left[0,~ 100\right]$\\
\hline
 F-2 & 80 & $70.7 \pm 13.0$ $\left[24.4,~ 85.3\right]$&  $57.9  \pm 13.6$ $\left[13.9,~ 74.4\right]$&  $71.6  \pm 10.5$ $\left[34.8,~ 85.8\right]$& $74.4  \pm 15.1$  $\left[15.0,~ 90.9\right]$\\
  \hline
  \end{tabular}
}
  \caption{Examination of varying number of patient datasets using framework 1, in four metrics of Dice, JI, precision and recall. Mean, standard deviation, lower and upper performance ranges are reported. Comparison of the presented framework 1 (F-1) versus framework 2 (F-2) in 80 patients is also presented.}\label{F1Eval}
\end{table*}
%\efloatseparator

%\begin{table*}
% \centerline{
%  \begin{tabular}{ccccc}
%  \hline
%   &   SI (\%) &   JI (\%) &   Precision (\%) &   Recall (\%) \\ \hline
%  F-1 &  $68.8 \pm 25.6$ $\left[0,~ 96.6\right]$&   $57.2  \pm 25.4$ $\left[0,~ 93.5\right]$ &  $71.5  \pm 30.0$ $\left[0,~ 100\right]$ &  $72.5  \pm 27.2$ $\left[0,~ 100\right]$\\ \hline
%  F-2 &  $70.7 \pm 13.0$ $\left[24.4,~ 85.3\right]$&  $57.9  \pm 13.6$ $\left[13.9,~ 74.4\right]$&  $71.6  \pm 10.5$ $\left[34.8,~ 85.8\right]$& $74.4  \pm 15.1$  $\left[15.0,~ 90.9\right]$\\
%  \hline
%  \end{tabular}
%}
%  \caption{Comparison of the presented framework 1 (F-1) versus framework 2 (F-2) in 80 patients. Mean, standard deviation, lower and upper performance ranges are reported.}\label{F1vsF2Eval}
%\end{table*}
%\efloatseparator

We exploit two variations of pancreas segmentation in a perspective of bottom-up information propagation from image patches to (segments) superpixels. Both frameworks are carried-out in a six-fold cross validation (CV) manner. Our protocol is arguably harder than the ``leave-one-out'' (LOO) criterion in \cite{Wang,Wolz,Chu,Wolz12,Shimizu} since less patient datasets are used in training and more separate patient scans for testing. In fact, \cite{Wolz} does demonstrate a notable performance drop from using 149 patients in training versus 49 patients under LOO, i.e.,  the mean Dice coefficients decreased from $69.6\%\pm 16.7\%$ to $58.2\%\pm 20.0\%$. This indicates that the multi-atlas fusion approaches \cite{Wang,Okada,Wolz,Chu,Wolz12,Shimizu} may actually achieve lower segmentation accuracies than reported, if under the six-fold cross validation protocol. At 40 patients, our result using framework 1 is $2.2\%$ better than the reported results by \cite{Wolz} using 50 patients (Dice coefficients of $60.4\%$ versus $58.2\%$). Comparing to the usage of $N-1$ patient datasets directly in the memory for multi-atlas registration methods, our learned models are more compactly encoded into a series of patch- and superpixel-level random forest classifiers and the CNN classifier for patch labeling. The computational efficiency also has been drastically improved in the order of $6\sim8$ minutes per testing case (using a mix of Matlab and C implementation, $\sim50\%$ time for superpixel generation), compared to others requiring $10$ hours or more. The segmentation framework (F-2) using deep patch labeling confidences is also more numerically stable, with no complete failure case and noticeable lower standard deviations.

\textcolor{LightRubineRed}
{{\bf Comparison to R-CNN and its variations \cite{Roth15,Roth2015}:} The conventional approach for classifying superpixels or image segments in computer vision is ``bag-of-words'' \cite{Carreira12b,Chatfield11}. ``Bag-of-words'' methods compute dense SIFT, HOG and LBP image descriptors, embed these descriptors through various feature encoding schemes and pool the features inside each superpixel for classification. Both model complexity and computational expense \cite{Carreira12b,Chatfield11} are very high, comparing with ours (Sec. \ref{subsec:Seg}). Recently, a ``Regional CNN'' (R-CNN) \cite{Girshick14,Girshick2015} method is proposed and shows substantial performance gains in PASCAL VOC object detection and semantic segmentation benchmarks \cite{Everingham2015}, compared to previous ``Bag-of-words'' models.  A simple R-CNN implementation on pancreas segmentation has been explored in our previous work \cite{Roth15} which reports evidently worse result (Dice coefficient $62.9\% \pm16.1\%$) than our F-2 framework  (Dice $70.7 \pm 13.0\%$) that spatially pools the CNN patch classification confidences per superpixel. Note that R-CNN \cite{Girshick14,Girshick2015} is not an ''end-to-end'' trainable deep learning system: R-CNN first uses the pre-trained or fine-tuned CNNs as image feature extractors for superpixels and then the computed deep image features are classified by support vector machine models.}

\textcolor{LightRubineRed} 
{Our recent work \cite{Roth2015} is an extended version of pancreas segmentation from the region-based convolutional neural networks (R-CNN) for semantic image segmentation \cite{Girshick2015,Everingham2015}. In \cite{Roth2015}, 1) we exploit multi-level deep convolutional networks which sample a set of bounding boxes covering each image superpixel at multiple spatial scales in a “zoom-out” fashion \cite{Mostajabi2015}; 2) the best performing model in \cite{Roth2015} is a stacked $R^2$-ConvNet which operates in the joint space of CT intensities and the $Patch$-ConvNet dense probability maps, similar to F-2. With the above two method extensions, \cite{Roth2015} reports the Dice coefficient of $71.8 \pm 10.7\%$ in four-fold cross-validation (which is slightly better than $70.7 \pm 13.0\%$ of F-2 using the same dataset). However, \cite{Roth2015} can not be directly trained and tested on the raw CT scans as in this paper, due to the data high-imbalance issue between pancreas and non-pancreas superpixels. There are overwhelmingly more negative instances than positive ones if training the CNN models directly on all image superpixels from abdominal CT scans. Therefore, given an input abdomen CT, an initial set of superpixel regions is first generated or filtered by a coarse cascading process of operating the random forests based pancreas segmentation \cite{Farag2014} (similar to F-1), at low or conservative classification thresholds. Over $96\%$ original volumetric abdominal CT scan space has been rejected for the next step (see Fig. \ref{fig:ROCPatch} {\bf Left}). For pancreas segmentation, these pre-labeled superpixels serve as regional candidates with high sensitivity ($>$97$\%$) but low precision (generally called Candidate Generation or CG process). The resulting initial DSC is $27\%$ on average. Then \cite{Roth2015} evaluates several variations of CNNs for segmentation refinement (or pruning). F-2 performs comparably to the extended R-CNN version for pancreas segmentation \cite{Roth2015} and is able to run without using F-1 to generate pre-selected superpixel candidates (which nevertheless is required by \cite{Roth15,Roth2015}). As discussed above, we would argue that these hybrid approaches combining or integrating deep and non-deep learning components (like this work and \cite{Roth2015,Roth15,Girshick14,Girshick2015,Chen2015Semantic}) will co-exist with the other fully ``end-to-end'' trainable CNN systems \cite{Ronneberger2015,Long2015Fully} that may produce comparable or even inferior segmentation accuracy levels. For example, \cite{Chen2015Semantic} is a two-staged method of deep CNN image labeling followed by fully connected Conditional Random Field (CRF) post-optimization \cite{Philipp2011Efficient}, achieving $71.6\%$ intersection-over-union value versus $62.2\%$ in \cite{Long2015Fully}, on PASCAL VOC 2012 test set for semantic segmentation task \cite{Everingham2015}.}

%%%%%%%%%%%%%%%%%%%%
\begin{table*}
  \centerline{
  \begin{tabular}{cccccc}
  \hline
  Reference &   N &   SI (\%) &   JI (\%) &   Precision (\%) &   Recall (\%) \\ \hline
  \cite{Shimizu}&   20&   -&   57.9 &   -& -\\ \hline
	\cite{Okada} & 28 & - & 46.6 & - & - \\ \hline
  \cite{Wolz}&   150&   $69.6 \pm 16.7$ &  $55.5 \pm 17.1$ & $67.9 \pm 18.2$ &  $74.1 \pm 17.1$ \\ \hline
	\cite{Wolz}&   50&   $58.2 \pm 20.0$ &  $43.5 \pm 17.8$  &   - &   - \\ \hline
	\cite{Wolz12}&   100 &   65.5 &  49.6  &   70.7  &   62.9 \\ \hline
  \cite{Wang}&  100&   $65.5 \pm 18.6$ &   -&   -&   -\\ \hline
	\cite{Chu}&  100&   $69.1 \pm 15.3$ &   54.6 &   -&   -\\ \hline
	Framework 1 & 80&   $68.8 \pm 25.6$&   $57.2  \pm 25.4$&   $71.5  \pm 30.0$&   $72.5  \pm 27.2$\\ \hline
  Framework 2 & 80&   $70.7 \pm 13.0$&   $57.9  \pm 13.6$&   $71.6  \pm 10.5$&   $74.4  \pm 15.1$\\ \hline
	\textcolor{LightRubineRed} {MALF} & 80&   \textcolor{LightRubineRed}{$52.5 \pm 20.8$} &   \textcolor{LightRubineRed}{$38.1 \pm 18.3$} &   \textcolor{LightRubineRed}{-} &   \textcolor{LightRubineRed}{-} \\
\hline
  \end{tabular}
}
  \caption{Comparison of F-1 and F-2 in six-fold cross validation to the recent state-of-the-art methods \cite{Wang,Wolz,Chu,Wolz12,Shimizu,Okada} in LOO \textcolor{LightRubineRed}{and our implementation of ``multi-atlas and label fusion'' (MALF) using publicly available C++ code bases \cite{Modat2010,Wang2012} under the same six-fold cross validation. The proposed bottom-up pancreas segmentation methods of F-1 and F-2 significantly outperform their MALF counterpart:  $68.8 \pm 25.6\%$  (F-1), $70.7 \pm 13.0\%$ (F-2) versus $52.51 \pm 20.84\%$ in Dice coefficients (mean$\pm$std).}. %Numerical results are directly copied from the publications \cite{Wang,Wolz,Chu,Wolz12,Shimizu,Okada} where we choose the best result from different parameter configurations in \cite{Chu}.
	}\label{Table:LitComp}
\end{table*}
%%%%%%%%%%%%%%%%%%%%%%%%%

\textcolor{LightRubineRed}
{{\bf Comparison to MALF (under six-fold CV):} For the ease of comparison to the previously well studied ``multi-atlas and label fusion'' (MALF) approaches, we implement a MALF solution for pancreas segmentation using the publicly available C++ code bases \cite{Modat2010,Wang2012}. The performance evaluation criterion is the same {\bf six-fold patient splits for cross validation}, not the ``leave-one-patient-out'' (LOO) in \cite{Wang,Wolz,Chu,Wolz12,Shimizu,Okada}. Specifically, each atlas in the training folds is registered to every target CT image in the testing fold, by the fast free-form deformation algorithm developed in NiftyReg \cite{Modat2010}. Cubic B-Splines are used to deform a source image to optimize an objective function based on the normalized mutual information and a bending energy term. Grid spacing along three axes are set as 5 mm. The weight of the bending energy term is 0.005 and the normalized mutual information with 64 bins are used. The optimization is performed in three coarse-to-fine levels and the maximal number of iterations per level is 300. More details can be found in \cite{Modat2010}. The registrations are used to warp the pancreas in the atlas set (66, or 67 atlases) to the target image. Nearest-neighbor interpolation is employed since the labels are binary images. For each voxel in the target image, each atlas provided an opinion about the label. The probability of pancreas at any voxel $x$ in the target $U$ was determined by $\hat{L}(x) = \sum_{i=1}^{n} \omega_i(x) L_i(x)$ where $L_i(x)$ is the warped $i$-th pancreas atlas and $\omega_i(x)$ is a weight assigned to the $i$-th atlas at location $x$ with $\sum_{i=1}^{n} \omega_i(x) =1$; and n is the number of atlases. In our 6-fold cross validation experiments $n=66$ or $67$. We adopt the joint label fusion algorithm \cite{Wang2012}, which estimates voting weights $\omega_i(x)$ by simultaneously considering the pairwise atlas correlations and local image appearance similarities at $x$. More details about how to capture the probability that different atlases produce the same label error at location $x$ via a formulation of dependency matrix can be found in \cite{Wang2012}. The final binary pancreas segmentation label or map $L(x)$ in target can be computed by thresholding on $\hat{L}(x)$. The resulted MALF segmentation accuracy in Dice coefficients are $52.51 \pm 20.84\%$ in the range of  $\left[0\%,~ 80.56\%\right]$. This pancreas segmentation accuracy is noticeably lower than the mean Dice scores of $58.2\%\sim 69.6\%$ reported in \cite{Wang,Wolz,Chu,Wolz12,Shimizu,Okada} under the protocol of ``leave-one-patient-out'' (LOO) for MALF methods. This observation may indicate the performance deterioration of MALF from LOO (equivalent to 80-fold CV) to 6-fold CV which is consistent with the finding that the segmentation accuracy drops from $69.6\%$ to $58.2\%$ when only 49 atlases are available instead of 149 \cite{Wolz}.}

\textcolor{LightRubineRed}
{Furthermore, we take about 33.5 days to fully conduct the six-fold MALF cross validation experiments using a Windows server; whereas the proposed bottom-up superpixel cascade approach finishes in $\sim9$ hours for 80 cases (6.7 minutes per patient scan on average). In summary, using the same dataset and under six-fold cross-validation, our bottom-up segmentation method significantly outperforms its MALF counterpart: $70.7 \pm 13.0\%$ versus $52.51 \pm 20.84\%$ in Dice coefficients, while being approximately 90 times faster. Converting our Matlab/C++ implementation into pure C++ should expect further $2\sim3$ times speed-up. }

%In summary, two variations of pancreas segmentation methods are presented and exploited in a perspective of bottom-up information propagation from image patches to segments. The SLIC superpixel generation approach provided the best overall pancreas organ-level boundary recall by partitioning each 2D CT axial slice into over-segmentation label maps of all patients. Random forest classifier and cascade of RF classifiers were trained at the image patch, deep patch and superpixel-level respectively, via extracting multi-channel features.

%\textcolor{mygray}

\section{CONCLUSION AND DISCUSSION}

\textcolor{LightRubineRed}
{In this paper, we present a fully-automated bottom-up approach for pancreas segmentation in abdominal computed tomography (CT) scans. The proposed method is based on a hierarchical cascade of information propagation by classifying image patches at different resolutions and multi-channel feature information pooling at (segments) superpixels. Our algorithm flow is a sequential process of decomposing CT slice images as a set of disjoint boundary-preserving superpixels; computing pancreas class probability maps via dense patch labeling; classifying superpixels via aggregating both intensity and probability information to form image features that are fed into the cascaded random forests; and enforcing a simple spatial connectivity based post-processing. The dense image patch labeling can be realized by efficient random forest classifier on hand-crafted image histogram, location and texture features; or deep convolutional neural network classification on larger image windows (i.e., with more spatial contexts).}

The main component of our method is to classify superpixels into either pancreas or non-pancreas class. Cascaded random forest classifiers are formulated for this task and performed on the pooled superpixel statistical features from intensity values and supervisedly learned class probabilities ($P^{RF}$ and/or $P^{CNN}$). The learned class probability maps (e.g., $P^{RF}$ and $P^{CNN}$) are treated as the supervised semantic class image embeddings which can be implemented, via an open framework by various methods, to learn the per-pixel class probability response.

To overcome the low image boundary contrast issue in superpixel generation, which is however common in medical imaging, we suggest that efficient supervised edge learning techniques may be utilized to artificially ``enhance'' the strength of semantic object-level boundary curves in 2D or surface in 3D. For example, one of the future directions is to couple or integrate the structured random forests based edge detection \cite{Dollar15} into a new image segmentation framework (MCG: Multiscale Combinatorial Grouping) \cite{Arbelaez} which permits a user-customized image gradient map. This new approach may be capable to generate image superpixels that can preserve even very weak semantic object boundaries well (in the image gradient sense) and subsequently prevent segmentation leakage.

Finally, voxel-level pancreas segmentation masks can be propagated from the stacked superpixel-level classifications and further improved by an efficient boundary refinement post-processing, such as the narrow-band level-set based curve/surface evolution \cite{Shi08,Kohlberger}, or the learned intensity model based graph-cut \cite{Wolz}. Further examination into the sub-connectivity processes for the pancreas segmentation framework that considers the spatial relationships of splenic, portal and superior mesenteric veins with pancreas may be needed for future work.

%{Evaluating superpixels in PASCAL semantic segmentation challenge \cite{Everingham,Yadollahpour}.} %An upper-bound segmentation performance is exploited to evaluate the selection of superpixel methods and parameter configurations. This is referred as the {\em oracle} accuracy on Dice, or Jaccard Index (same as ``intersection over union'' or IOU in PASCAL).
\section*{ACKNOWLEDGEMENT}

This research was supported by the Intramural Research Program of the National Institutes of Health Clinical Center. We thank Andrew Dwyer, MD at National Institutes of Health Clinical Center for critical review of the manuscript.

%{
%\bibliographystyle{alpha}
%\bibliography{mybibfile}
%}

%\def\bibsection{\section*{References}}
%\bibliographystyle{model2-names}
%\bibliography{mybibfile}

\bibliographystyle{IEEEtran}
%\fontsize{8}{9}\selectfont
\bibliography{mybibfile}

\end{document}